\documentclass{l4dc2027}

\usepackage{times}
\usepackage{amsfonts}
\usepackage{textcomp}
\usepackage{array}
\usepackage[normalem]{ulem}
\usepackage{dsfont}
\usepackage{booktabs}
\usepackage{dashrule}
\usepackage{xcolor}
\usepackage{xspace}

\usepackage{enumitem}
\graphicspath{{figures/}}

\makeatletter
\let\l@origNewtheorem\newtheorem
\renewcommand{\newtheorem}[2]{\@ifundefined{#1}{\l@origNewtheorem{#1}{#2}}{}}

\def\R{\mathbb{R}}

\def\notes#1{\marginpar{\tiny #1}\typeout{Notes!
Notes!
Notes!
}}
\renewcommand{\notes}[1]{\typeout{notes!}}

\def\Expect{{\mathbb E}}

\def\R{\mathbb{R}}

\def\N{\mathcal{N}}

\newtheorem{proposition}{Proposition}

\def\beq{\begin{eqnarray}} 
\def\bc{\begin{center}} 
\def\be{\begin{enumerate}}
\def\bi{\begin{itemize}} 
\def\bs{\begin{small}}
\def\bS{\begin{slide}}
\def\ec{\end{center}} 
\def\ee{\end{enumerate}}
\def\ei{\end{itemize}}
\def\es{\end{small}}
\def\eS{\end{slide}}
\def\eeq{\end{eqnarray}}

\newcommand{\ud}{\,\mathrm{d}}

\newcounter{rmnum}

\newcounter{anum}

\let\newtheorem\l@origNewtheorem
\makeatother

\newif\iftodo \todofalse
\iftodo
  \newcommand{\todo}[1]{\textcolor{red}{[TODO: #1]}}
  \newcommand{\tryexp}[1]{\textcolor{blue}{[EXPERIMENT: #1]}}
  \newcommand{\suggest}[1]{\textcolor{magenta}{[SUGGESTION: #1]}}
  \newcommand{\tbd}[1]{\textcolor{red}{\fbox{#1}}}
  
  \newenvironment{addedblock}{\color{green!50!black}}{}
\else
  \newcommand{\todo}[1]{}
  \newcommand{\tryexp}[1]{}
  \newcommand{\suggest}[1]{}
  \newcommand{\tbd}[1]{??}

\fi

\newcommand{\OTP}{OTP\xspace}                              %
\newcommand{\FMP}{FMP\xspace}                              %
\newcommand{\SIP}{SIP\xspace}                              %
\newcommand{\KRP}{KRP\xspace}                              %
\newcommand{\DP}{DP\xspace}                                          %
\newcommand{\CP}{CP\xspace}                                          %
\newcommand{\DPR}{DPR\xspace}                                        %
\newcommand{\OTPlong}{optimal transport policy}
\newcommand{\FMPlong}{flow matching policy}
\newcommand{\SIPlong}{stochastic interpolant policy}
\newcommand{\KRPlong}{Knothe--Rosenblatt policy}
\newcommand{\OTPLong}{Optimal transport policy}
\newcommand{\FMPLong}{Flow matching policy}
\newcommand{\SIPLong}{Stochastic interpolant policy}
\newcommand{\KRPLong}{Knothe--Rosenblatt policy}

\makeatletter
\providecommand{\captionof}[1]{\def\@captype{#1}\caption}
\newcommand{\narrowcaptions}{\long\def\@makecaption##1##2{\vskip 3pt\noindent ##1: ##2\par\vskip 3pt}}
\makeatother

\title[Conditional Optimal Transport and Triangular Maps for Imitation Learning]{Conditional Optimal Transport and Triangular Maps \\ for Imitation Learning}

\author{%
    \Name{Mohammad Al-Jarrah} \Email{mohd9485@uw.edu}\\
    \addr Department of Aeronautics \& Astronautics and Department of Applied Mathematics,\\
     University of Washington, Seattle, WA, USA%
    \AND
    \Name{Yuhang Mei} \Email{yuhangm@uw.edu}\\
    \addr Department of Aeronautics and Astronautics, University of Washington, Seattle, WA
    \AND
    \Name{Karen Leung} \Email{kymleung@uw.edu}\\
    \addr Department of Aeronautics and Astronautics, University of Washington, Seattle, WA
    \AND
    \Name{Bamdad Hosseini} \Email{bamdadh@uw.edu}\\
    \addr Department of Applied Mathematics,  University of Washington, Seattle, WA%
    \AND
    \Name{Amirhossein Taghvaei} \Email{amirtag@uw.edu}\\
    \addr Department of Aeronautics and Astronautics, University of Washington, Seattle, WA
}

\begin{document}

\maketitle

\begin{abstract}%
We introduce a unified formulation of imitation learning from demonstrations as a conditional sampling problem solved by triangular transport: a policy is the second block of a triangular map that transports a reference distribution to the conditional distribution of the action chunk given the observation history. Unlike in Diffusion Policy, where it is a fixed Gaussian, the reference becomes a design variable; we take it to be the empirical distribution of the demonstrated action chunks and introduce three instances of the formulation: an \OTPlong{} (\OTP), a \FMPlong{} (\FMP), and a \SIPlong{} (\SIP). On the Push-T and robomimic Square tasks, the three policies are competitive in task performance with Diffusion Policy and Consistency Policy while using considerably smaller networks and shorter training, and the informative reference accounts for a substantial part of their performance when few network evaluations are used. The three instances offer different trade-offs rather than a single best method: \OTP generates an action chunk with a single evaluation of the smallest network ($0.23$ million parameters on Push-T) and is suitable for settings in which inference time and model size matter, whereas \SIP attains the highest scores among the proposed policies at a higher cost.
\end{abstract}

\begin{keywords}%
Imitation Learning, Conditional Sampling, Generative Models, Optimal Transport
\end{keywords}

\section{Introduction}\label{sec:intro}

Imitation learning aims to compute a control policy that reproduces the behavior of a demonstrator. Its most widely used form, behavior cloning, casts the problem as supervised learning, with recorded observations as inputs and demonstrated actions as targets~\citep{mandlekar2021matters}. The conditional action distribution, however, can be multimodal, with several distinct actions being valid under the same observation~\citep{florence2022implicit,shafiullah2022behavior,MizutaLeung2026,SchmerlingLeungEtAl2018}. 
This diversity motivates policies that represent the entire conditional distribution of the action given the observation, so that the problem is one of conditional sampling: given only joint samples of observations and actions, generate actions distributed according to their conditional distribution given a new observation, fast enough to be executed online.

\begin{figure}[t]
\centering
\includegraphics[width=0.66\textwidth]{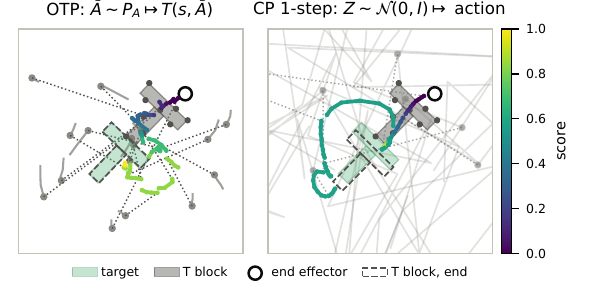}
\caption{One Push-T episode under \OTP (left) and \CP~1-step (right) from the same observation $s$, up to step $61$, at which \OTP completes the task. Every action chunk is the image of a fresh sample of the reference, $P_A$ for \OTP and $\N(0, I_n)$ for \CP (gray; about half of the points of the latter lie outside the frame), and dotted lines connect the first action of each sample to that of its image. The executed actions are colored by the score after their execution, and \CP completes this episode at step $265$ (see Appendix~\ref{app:eval} for details). 
}
\label{fig:overview}
\end{figure}

\emph{Gaussian-reference policies.} Diffusion Policy (\DP)~\citep{chi2023diffusion} represents 
the conditional distribution of the action
by a 
diffusion model~\citep{ho2020denoising} conditioned on the recent observations, and it has become a strong baseline for manipulation tasks. Its sampler, however, generates every chunk of future actions by iteratively denoising a Gaussian sample, at the cost of one evaluation of a large network per 
integration step: $100$ steps in its simulation benchmarks, reduced to $10$ by deterministic sampler~\citep{song2021denoising} in its real-world experiments. This cost has motivated samplers with fewer steps~\citep{zhang2023fast}, distilled policies such as Consistency Policy (\CP)~\citep{prasad2024consistency}, and flow-matching policies (see Appendix~\ref{app:related} for details). All of these start from a Gaussian distribution, which carries no information about the task.

\emph{Informative references.} A growing body of work replaces the Gaussian reference with a more informative distribution, which we simply refer to as \emph{the reference}. 
The adaptation of the reference distribution has already been considered in
BRIDGeR~\citep{chen2024behavioral}, which interpolates from a learned or hand-crafted reference policy and reports gains at a small number of integration steps, as well as other methods that construct the reference from previously executed actions~\citep{kang2026warmprior,jiang2025streaming} or learn it as a function of the observation~\citep{dai2026where,liu2026sample}. All of these approaches, however, require a separately trained model, domain knowledge, or the pairing of every training target with its own historical data (see Appendix~\ref{app:related} for details). This raises the question of whether a reference can be informative and yet independent of both the observation and the previously executed actions, which would lead to more flexible and efficient algorithms.

In this paper, we build upon the generative filters of \citet{aljarrah2026generative}, in which the analysis step of nonlinear filtering is realized by a generative model that transports the forecast distribution of the state to its posterior. The problem solved by a policy, namely sampling from the conditional distribution of the action given the observation using only joint samples of the two, coincides with this conditioning problem. Following a triangular transport formulation~\citep{kovachki2020conditional,taghvaei2022optimal,al-jarrah2024nonlinear}, we represent a policy as the second block of a triangular map, and we take the reference to be the empirical distribution of the demonstrated action chunks, so that every action chunk is obtained by transporting one that the demonstrator executed. To our knowledge, no existing policy takes as reference the empirical distribution of the demonstrated action chunks itself (see Appendix~\ref{app:related} for details). Three policies are derived in this manner, each from one of the filters of \citet{aljarrah2026generative}: the optimal transport filter (OTF)~\citep{al-jarrah2024nonlinear} yields the \OTPlong{} (\OTP), which requires a single network evaluation per action chunk; the flow matching filter (FMF) yields the \FMPlong{} (\FMP), realized by an ordinary differential equation (ODE)~\citep{lipman2023flow}; and the stochastic interpolant filter (SIF) yields the \SIPlong{} (\SIP), realized by a stochastic differential equation (SDE)~\citep{albergo2025}. In addition, the stochastic map filter of \citet{spantini2022coupling}, which is based on the Knothe--Rosenblatt rearrangement and is considered in \citet{aljarrah2026generative}, yields the \KRPlong{} (\KRP), which we include as a Gaussian-reference counterpart of the proposed policies. We refer to these methods as policies rather than filters, since each conditions the action chunk on the observation history at a single re-planning instant, without the recursive prediction and update in filters.

The contributions of this paper are as follows.
\begin{enumerate}[leftmargin=*, itemsep=0.2ex, topsep=0.4ex]
\item \emph{Formulation.} We formulate imitation learning with action chunks as a conditional sampling problem and represent a policy as the second block of a triangular transport map, so that the reference distribution becomes a design variable and \DP and \CP are recovered as the instances with a Gaussian reference (Section~\ref{sec:triangular} and Appendix~\ref{app:baselines}).
\item \emph{Policies, reference, and deployment.} Within this formulation, we derive three policies, \OTP{}, \FMP{}, and \SIP, which learn the map by conditional optimal transport (OT), flow matching, and stochastic interpolants, respectively, with the reference $P_A$, which requires no auxiliary model and no pairing of the training data, together with a deployment protocol under which the reference has the same distribution during training and deployment (Section~\ref{sec:methods}); \OTP is learned in a single training stage and evaluates, on Push-T, a network of $0.23$ million parameters once per action chunk. 
\item \emph{Evaluation.} We compare the proposed policies with \KRP{}, \DP{}, and \CP under a common protocol, on the Push-T benchmark and on the Square task of robomimic, in terms of task performance and of training and online cost (Section~\ref{sec:numerics}), and we show that the reference $P_A$ is responsible for a substantial part of their performance, in particular with few network evaluations, the regime of interest for online deployment.
\end{enumerate}

The remainder of this paper is organized as follows. Section~\ref{sec:problem} formulates imitation learning with action chunks as a conditional sampling problem. Section~\ref{sec:methods} presents the triangular formulation, the proposed policies, and the deployment protocol. Section~\ref{sec:numerics} compares the proposed policies with the Gaussian-reference policies on Push-T and Square, and Section~\ref{sec:discussion} discusses the results, the limitations, and future work. 
The appendices provide supplementary material.

\section{Problem Formulation}\label{sec:problem}

Consider a system operated by a demonstrator at the discrete time instants $k = 0, 1, 2, \ldots$, where $o_k \in \R^{d_o}$ denotes the observation and $a_k \in \R^{d_a}$ the action applied at time $k$. A demonstration data set $\mathcal{D} = \{(o^j_k, a^j_k) : k = 0,\ldots,K_j - 1,\ j = 1,\ldots,J\}$ consists of $J$ episodes, the $j$-th of which has length $K_j$. Following \citet{zhao2023learning,chi2023diffusion}, the policy returns a chunk of future actions conditioned on a short history of observations; to this end, we define the \emph{observation history} and the \emph{action chunk}
\begin{equation}\label{eq:chunks}
  S_k := (o_{k-H_o+1}, \ldots, o_k) \in \R^{m}, \qquad
  A_k := (a_k, \ldots, a_{k+H_p-1}) \in \R^{n},
\end{equation}
where $H_o$ is the observation horizon, $H_p$ is the prediction horizon, $m = H_o d_o$, and $n = H_p d_a$. We take $S_k$ to consist of observations only, since a policy conditioned on its own past actions tends to copy them rather than respond to the observations~\citep{wen2020fighting,dehaan2019causal}. Collecting the pairs~\eqref{eq:chunks} from the demonstration data set $\mathcal D$ yields the \emph{training data} $\{(S^i, A^i)\}_{i=1}^{M}$, where $M$ denotes the number of available training data pairs. We regard the training data as samples of the joint distribution $P_{S,A}$ of the observation history and the action chunk under the demonstrator, with marginals $P_S$ and $P_A$; every expectation with respect to these distributions is approximated by an average over the training data, and samples of $P_A$ are drawn uniformly from the $M$ demonstrated action chunks. At deployment, the policy is executed in a receding-horizon fashion: at a re-planning instant $k$, an action chunk $\hat A_k = (\hat a_k, \ldots, \hat a_{k+H_p-1})$ is sampled given $S_k$, its first $H_a \le H_p$ actions are applied, and the next chunk is sampled at time $k + H_a$. At the first instants of an episode, $k < H_o - 1$, the missing observations of $S_k$ are replaced by the first observation $o_0$, as in the training data. The horizons $H_o$, $H_p$, and $H_a$ are selected separately for each policy (Section~\ref{sec:tuning}).

\noindent\textbf{Problem statement.}
Let $\{(S^i, A^i)\}_{i=1}^{M}$ be the training data, and let $P_Z$ be a \emph{reference} distribution on $\R^{n}$ from which samples can be drawn. The goal is to construct a map $T$ that reproduces the conditional distribution of the action chunk given the observation history:
\begin{align*}
  \text{Given:}\quad & \bigl(S^i,\,A^i\bigr)_{i=1}^{M}, \quad (S^i, A^i) \sim P_{S,A}, \\
  \text{Find:}\quad & \textit{a map }
    T : \R^{m}\times\R^{n} \to \R^{n} \textit{ such that }
    T(s,\cdot)_{\#}\,P_Z = P_{A \mid S = s}
    \textit{ for $P_S$-a.e.\ } s \in \R^{m},
\end{align*}
where $\#$ denotes the push-forward operator. A solution defines a stochastic policy, in which the action chunk $\hat A_k = T(S_k, Z_k)$ is computed from a reference sample $Z_k \sim P_Z$ drawn afresh, independently of $S_k$, at every re-planning instant.

This problem coincides with the conditioning problem of the analysis step of nonlinear filtering~\citep{aljarrah2026generative}, with the observation history in place of the observation and the action chunk in place of the state, except that the map is learned once, offline, from recorded demonstrations, and that, at deployment, the observation history is generated in closed loop and may leave the support of $P_S$, a phenomenon known as covariate shift~\citep{ross2011reduction}. Consequently, a policy is assessed by the closed-loop performance on the task (Section~\ref{sec:numerics}) rather than by a distance between conditional distributions. 

\section{Generative Transport Policies}\label{sec:methods}

\subsection{Conditioning with triangular maps}\label{sec:triangular}

The policies studied in this paper share one element: each is realized by a \emph{triangular} map, learned from the training data by a generative modeling framework. We say that a map $G : \R^{m}\times\R^{n} \to \R^{m}\times\R^{n}$ is triangular if there exists a map $T : \R^{m}\times\R^{n} \to \R^{n}$ such that
\begin{equation}\label{eq:triangular-map}
  G(s, z) = \bigl(s,\, T(s, z)\bigr), \qquad \forall\, (s, z) \in \R^{m}\times\R^{n},
\end{equation}
that is, $G$ holds the observation history fixed. For some of the policies, such as the stochastic samplers of Appendices~\ref{sec:sif} and~\ref{sec:dp}, the map  $T$ depends on auxiliary randomness that is independent of $(s, z)$.  In such cases, without loss of generality, the reference is extended by the auxiliary variable. Triangular maps are useful for conditioning because of the following result~\citep{kovachki2020conditional,al-jarrah2024nonlinear}, whose short proof is given in Appendix~\ref{app:proof}.

\begin{proposition}\label{prop:triangular}
Let $P_{S,A}$ be a probability distribution on $\R^{m}\times\R^{n}$ with marginal $P_S$ on $\R^{m}$, let $P_Z$ be a probability distribution on $\R^{n}$, and let $G$ be a measurable triangular map of the form~\eqref{eq:triangular-map}. If $G_{\#}(P_S \otimes P_Z) = P_{S,A}$, then $T(s,\cdot)_{\#} P_Z = P_{A \mid S = s}$ for $P_S$-almost every $s$, where $P_{A \mid S = s}$ denotes the conditional distribution obtained by disintegrating $P_{S,A}$ with respect to $P_S$.
\end{proposition}

This result suggests a recipe for converting a generative model into a policy~\citep{aljarrah2026generative}: (i)~let the \emph{target} be the joint distribution $P_{S,A}$ of the training data; (ii)~let the \emph{source} be the product distribution $P_S \otimes P_Z$, whose samples are obtained by pairing every observation history with a reference sample drawn independently of it; and (iii)~train a generative model to learn a (deterministic or stochastic) map of the triangular form~\eqref{eq:triangular-map} that transports the source to the target. Any such map satisfies the hypothesis of Proposition~\ref{prop:triangular}, and its second block consequently solves the problem of Section~\ref{sec:problem}. This condition does not specify $T$ uniquely, and the policies presented below differ precisely in how a particular solution is selected and learned.

The reference $P_Z$ is a design choice that Proposition~\ref{prop:triangular} leaves entirely free: \DP{}, \CP{}, and the flow-matching and distilled policies take $P_Z = \N(0, I_n)$, whereas the informative references of Section~\ref{sec:intro} depend on the observation or on the previously executed actions, so that the source is no longer of the form $P_S \otimes P_Z$. We instead take $P_Z = P_A$, so that the policy reads
\begin{equation}\label{eq:policy}
  \hat A_k = T\bigl(S_k,\, \bar A_k\bigr), \qquad \bar A_k \sim P_A,
\end{equation}
where $\bar A_k$ is drawn independently of $S_k$ at every re-planning instant. This reference is supported on the actions applied by the demonstrator, which are confined to the reachable workspace and vary smoothly in time, properties that a Gaussian does not possess (Figure~\ref{fig:overview}). Moreover, since $P_A$ is the mixture of the conditional distributions $P_{A \mid S = s}$ over $s \sim P_S$, the transport to $P_{A \mid S = s}$ is expected to require a smaller displacement from $P_A$ than from a Gaussian, which would simplify the learning of the map and reduce the number of integration steps, consistently with the observations of \citet{chen2024behavioral,kang2026warmprior} for informative references. For \OTP and \FMP, whose maps are deterministic, the randomness of the policy is thus inherited entirely from the demonstrations, and, for \SIP, it is supplemented by the noise of the sampler. 

\subsection{The proposed policies}\label{sec:proposed}

Following \citet{aljarrah2026generative}, we obtain three policies with the reference $P_A$, each of which realizes the map $T$ of~\eqref{eq:policy} by the generative modeling framework of one of the filters of that work (see Appendix~\ref{app:derivation} for details). As summarized in Table~\ref{tab:methods}, they differ in the optimization problem by which $T$ is learned, in the form of the transport at deployment, and in the randomness of the policy at a fixed reference sample: \OTP{}, derived from the optimal transport filter~\citep{al-jarrah2024nonlinear}, learns the conditional OT map from a max--min problem and transports the reference by a one-shot map; \FMP{}, derived from the flow matching filter, learns a vector field by least-squares regression~\citep{lipman2023flow} and transports the reference along a deterministic flow, an ODE; and \SIP{}, derived from the stochastic interpolant filter, learns a vector field and a score by two least-squares regressions~\citep{albergo2025} and transports the reference along a stochastic flow, an SDE. Consequently, \OTP trades the stability of a regression objective for an online cost of a single network evaluation, whereas \FMP and \SIP are trained by regression at the price of $200$ network evaluations; we note that \SIP is formally the instance of BRIDGeR~\citep{chen2024behavioral} whose source policy is the action marginal.

\subsection{Deployment protocol}\label{sec:deployment}

The deployment of the proposed policies is summarized by Algorithm~\ref{alg:deploy} in Appendix~\ref{app:impl}. The reference sample $\bar A_k$ is drawn afresh from the demonstrations at every re-planning instant, independently of the observation history, so that the reference has the same distribution during training and deployment. Reusing instead the action chunk produced at the previous re-planning instant violates this independence, and it was markedly inferior in preliminary experiments.

\section{Numerical Results}\label{sec:numerics}\label{sec:gaussref}

In this section, we list the policies that are compared, describe the benchmarks and the evaluation protocol, and present the results, with an emphasis on the trade-off between task performance and computational cost. The code that reproduces all the experiments is publicly available at \url{https://github.com/Mohd9485/transport-policies}.

\subsection{Policies compared}\label{sec:compared}

Table~\ref{tab:methods} lists the policies, which fall into two groups. The first comprises the proposed policies \OTP{}, \FMP{}, and \SIP{} of Section~\ref{sec:proposed}, which transport the reference $P_A$. The second comprises four policies that transport the Gaussian reference $\N(0, I_n)$, each of which is an instance of the formulation of Section~\ref{sec:triangular}: \KRP{} learns the triangular map in the inverse direction, from the data to the reference, by maximum likelihood, and inverts it coordinate by coordinate at deployment; \DP~\citep{chi2023diffusion} generates an action chunk by $100$ steps of an ancestral sampler; \DPR (Diffusion Policy with a residual network) trains the objective of \DP on the residual network and the integrator of \FMP and \SIP{}, so that its comparison with \SIP separates the effect of the reference from that of the network; and \CP~\citep{prasad2024consistency} distills a diffusion teacher into a student that is evaluated once (\CP~1-step) or three times (\CP~3-step; see Appendix~\ref{app:baselines} for details). Two variations are, in addition, examined. On Push-T, \OTP{}, \FMP{}, and \SIP are trained with the reference $\N(0, I_n)$ in place of $P_A$, and \FMP{}, \SIP{}, and \DPR are deployed with fewer integration steps. On Square, every policy is evaluated without and with two modifications, the coordinate-wise mean of $8$ sampled chunks and Gaussian noise added to the state of the robot during training (Section~\ref{sec:square}).

\begin{table}[t]
\caption{Summary of the policies evaluated in this paper; related policies are classified in Table~\ref{tab:related}.}
\label{tab:methods}
\centering
\footnotesize
\setlength{\tabcolsep}{3pt}
\begin{tabular}{@{}>{\raggedright\arraybackslash}p{0.065\textwidth}
  >{\raggedright\arraybackslash}p{0.095\textwidth}
  >{\raggedright\arraybackslash}p{0.255\textwidth}
  >{\raggedright\arraybackslash}p{0.195\textwidth}
  >{\raggedright\arraybackslash}p{0.28\textwidth}@{}}
\toprule
Policy & Reference & Transport / coupling & Optimization & Transport at deployment (NFE) \\
\midrule
\OTP          & $P_A$ & OT map, independent coupling & max--min~\eqref{eq:maxmin} & one-shot map ($1$) \\
\FMP          & $P_A$ & deterministic interpolant~\eqref{eq:lininterp} & least squares~\eqref{eq:fmfloss} & deterministic flow, ODE~\eqref{eq:fmfode}, Heun ($200$) \\
\SIP      & $P_A$ & stochastic interpolant~\eqref{eq:sifinterp} & least squares~\eqref{eq:fmfloss}, \eqref{eq:sifden} & stochastic flow, SDE~\eqref{eq:sifsde}$^\dagger$, stochastic Heun ($200$) \\
\KRP & $\N(0, I_n)$ & Knothe--Rosenblatt map, inverse direction~\eqref{eq:krfmap} & maximum likelihood~\eqref{eq:krfloss} & inverse map, coordinate-wise Newton ($\le 192^{\ddagger}$) \\
\DP   & $\N(0, I_n)$ & noising process & denoising~\eqref{eq:ddpmloss} & stochastic chain, ancestral sampler$^\dagger$ ($100$) \\
\DPR & $\N(0, I_n)$ & noising process, continuous time & denoising~\eqref{eq:ddpmloss} & deterministic flow, probability-flow ODE, Heun ($200$) \\
\CP & $\N(0, I_n)$ & distilled from a diffusion teacher & consistency distillation & one-shot map ($1$) or three-step map$^\dagger$ ($3$) \\
\bottomrule
\end{tabular}
\par\vspace{0.4ex}
\parbox{\textwidth}{\scriptsize NFE: network evaluations per action chunk. $^\dagger$Stochastic inference: repeated evaluation at a fixed reference sample yields different action chunks. $^\ddagger$Evaluations of a single component of $\Phi_\theta$, each at $16$ quadrature nodes: at most $12$ per coordinate of the action chunk ($2$ to bracket the root and at most $10$ Newton steps), for $n = 16$, after one evaluation of the whole map that yields the reference sample.}
\end{table}

\subsection{Benchmarks and evaluation protocol}\label{sec:pusht}\label{sec:tuning}

We consider the state-based Push-T task~\citep{florence2022implicit,chi2023diffusion}, in which a circular end effector moving in a plane must push a T-shaped block from a randomized initial pose into a fixed target pose, with $a_k \in \R^{2}$ and $o_k \in \R^{20}$. The data set consists of $J = 206$ human demonstrations with $M = 25{,}650$ time instants in total. As in \citet{chi2023diffusion}, the reward is $r_k = \min\{c_k / 0.95,\, 1\}$, where $c_k \in [0, 1]$ is the fraction of the target region covered by the block at time $k$, an episode terminates once $c_k > 0.95$ or after $300$ steps, its \emph{score} is $\max_k r_k$, and it is called a \emph{success} if its score equals one; the score of a policy is its mean episode score. Every policy is trained with three training seeds on all $206$ demonstrations and evaluated, with the final weights of every training run and without any selection on the evaluation episodes, on the same $500$ held-out initial conditions, so that each policy is assessed on the same $1{,}500$ episodes. The hyperparameters of the proposed policies are listed in Table~\ref{tab:hyper}, the baselines are used with their published settings, and the horizons $(H_o, H_p, H_a)$ are $(1, 4, 4)$ for \OTP and \FMP, $(1, 8, 8)$ for \SIP and \KRP, and $(2, 16, 8)$ for \DP and \CP{} (see Appendices~\ref{app:impl} and~\ref{app:eval} for details).
Figure~\ref{fig:pushtepisodes} in Appendix~\ref{app:eval} shows the final configurations of nine held-out episodes for every policy.

We consider, in addition, the Square task of robomimic~\citep{mandlekar2021matters}, simulated in robosuite~\citep{zhu2020robosuite}, in which a robot arm must grasp a square nut and place it on a peg; the data set consists of the $200$ proficient-human demonstrations ($30{,}154$ time instants), with $o_k \in \R^{23}$ and $a_k \in \R^{10}$, and an episode is a success if the nut is placed within $400$ steps. The evaluation protocol is the same, and the two modifications of Section~\ref{sec:compared} were selected on a disjoint block of initial conditions. Figure~\ref{fig:squareepisodes} shows the final configurations of six held-out episodes for every policy, without and with the two modifications (see Appendix~\ref{app:square} for details).

\subsection{Results}\label{sec:results}\label{sec:cost}\label{sec:reference}\label{sec:square}

\begin{table}[t]
\caption{Task performance and computational cost on Push-T, over three training seeds and $500$ held-out initial conditions ($1{,}500$ episodes per policy).}
\label{tab:results}\label{tab:cost}
\centering
\footnotesize
\setlength{\tabcolsep}{2pt}
\begin{tabular}{@{}lcccccccc@{}}
\toprule
 & \multicolumn{4}{c}{Task performance} & \multicolumn{4}{c}{Computational cost} \\
\cmidrule(lr){2-5}\cmidrule(l){6-9}
Policy & Score$\uparrow$ & Seed spread$\downarrow$ & Success [\%]$\uparrow$ & Failure [\%]$\downarrow$ & NFE$\downarrow$ & Param.\ [M]$\downarrow$ & Training [min]$\downarrow$ & Time [s]$\downarrow$ \\
\midrule
\OTP       & $0.943 \pm 0.005$ & $0.016$ & $51.2$ & $4.7$ & $\mathbf{1}$ & $\mathbf{0.23}^{a}$ & $10.6$ & $\mathbf{0.12}$ \\
\FMP       & $0.947 \pm 0.004$ & $0.008$ & $52.1$ & $4.4$ & $200$ & $8.10$ & $\mathbf{0.6}$ & $3.74$ \\
\SIP       & $\mathbf{0.965} \pm 0.004$ & $0.005$ & $\mathbf{66.7}$ & $\mathbf{3.1}$ & $200$ & $2.67$ & $1.9$ & $2.10$ \\
\KRP       & $0.914 \pm 0.006$ & $0.018$ & $55.6$ & $8.2$ & $\le 192^{b}$ & $6.88$ & $7.3$ & $1.25$ \\
\DP        & $0.915 \pm 0.007$ & $0.004$ & $\mathbf{66.7}$ & $8.6$ & $100$ & $65.8$ & $268.3$ & $5.40$ \\
\DPR       & $0.963 \pm 0.004$ & $0.009$ & $56.6$ & $\mathbf{3.1}$ & $200$ & $8.47$ & $2.4$ & $2.84$ \\
\CP 1-step & $0.938 \pm 0.006$ & $0.011$ & $63.0$ & $5.9$ & $\mathbf{1}$ & $69.5^{c}$ & $309.6^{c}$ & $0.17$ \\
\CP 3-step & $0.940 \pm 0.006$ & $0.011$ & $62.7$ & $5.7$ & $3$ & \multicolumn{2}{c}{as \CP 1-step} & $0.29$ \\
\bottomrule
\end{tabular}
\par\vspace{0.4ex}
\parbox{\textwidth}{\scriptsize Score: mean episode score $\pm$ one standard error over the $1{,}500$ episodes. Seed spread: population standard deviation of the three per-seed mean scores. Success: score equal to one. Failure: score below $0.5$. NFE: network evaluations per action chunk. Param.: parameters of the network evaluated at inference. Training: wall-clock time of one training run, mean over the three training seeds, each run alone on one GPU. Time: wall-clock time of the evaluation divided by its $1{,}500$ episodes, including the simulation of the environment, a measure of throughput rather than of latency. The open-loop replay of the $206$ demonstrations scores $0.947$. $^{a}$The map $T$. $^{b}$Evaluations of a single component of the \KRP map, after one evaluation of the whole map that yields the reference sample (Table~\ref{tab:methods}). $^{c}$The student, distilled from a teacher of $65.8$ million parameters; the two variants of \CP share the trained networks. See Appendix~\ref{app:eval} for details.}
\end{table}

\begin{figure}[t]
\centering
\includegraphics[width=0.78\textwidth]{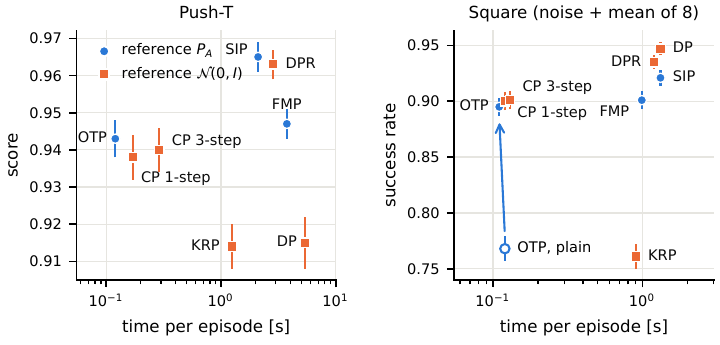}
\caption{Score (Push-T) and success rate (Square) against the time per episode, mean $\pm$ one standard error over $1{,}500$ episodes; on Square, every policy is trained with state noise and executes the mean of $8$ sampled chunks (Section~\ref{sec:square}), and the open circle marks \OTP without these modifications.}
\label{fig:scoretime}
\end{figure}

\begin{table}[t]
\caption{Success rate and computational cost on Square, over three training seeds and $500$ held-out initial conditions ($1{,}500$ episodes per entry), without and with the two modifications of Section~\ref{sec:compared}.}
\label{tab:square}
\centering
\footnotesize
\setlength{\tabcolsep}{2.3pt}
\begin{tabular}{@{}lccccccc@{}}
\toprule
 & \multicolumn{4}{c}{Success rate} & \multicolumn{3}{c}{Computational cost} \\
\cmidrule(lr){2-5}\cmidrule(l){6-8}
Policy & Plain$\uparrow$ & Mean of $8$$\uparrow$ & State noise$\uparrow$ & Both$\uparrow$ & Param.\ [M]$\downarrow$ & Training [min]$\downarrow$ & Time [s]$\downarrow$ \\
\midrule
\OTP       & $0.768 \pm 0.011$ & $0.871 \pm 0.009$ & $0.841 \pm 0.009$ & $0.895 \pm 0.008$ & $\mathbf{11.0}^{a}$ & $113.7$ & $\mathbf{0.12}$ \\
\FMP       & $0.856 \pm 0.009$ & $0.865 \pm 0.009$ & $0.907 \pm 0.008$ & $0.901 \pm 0.008$ & $12.8$ & $\mathbf{18.6}$ & $0.98$ \\
\SIP       & $0.913 \pm 0.007$ & $0.897 \pm 0.008$ & $0.936 \pm 0.006$ & $0.921 \pm 0.007$ & $24.1$ & $34.9$ & $1.24$ \\
\KRP       & $0.753 \pm 0.011$ & $0.721 \pm 0.012$ & $0.821 \pm 0.010$ & $0.761 \pm 0.011$ & $18.2$ & $61.1$ & $0.84$ \\
\DP        & $\mathbf{0.938} \pm 0.006$ & $\mathbf{0.939} \pm 0.006$ & $\mathbf{0.953} \pm 0.005$ & $\mathbf{0.947} \pm 0.006$ & $65.9$ & $322.3$ & $1.24$ \\
\DPR       & $0.896 \pm 0.008$ & $0.891 \pm 0.008$ & $0.942 \pm 0.006$ & $0.935 \pm 0.006$ & $18.2$ & $46.0$ & $1.05$ \\
\CP 1-step & $0.879 \pm 0.008$ & $0.880 \pm 0.008$ & $0.889 \pm 0.008$ & $0.900 \pm 0.008$ & $69.6^{c}$ & $317.0^{c}$ & $\mathbf{0.12}$ \\
\CP 3-step & $0.887 \pm 0.008$ & $0.881 \pm 0.008$ & $0.897 \pm 0.008$ & $0.901 \pm 0.008$ & \multicolumn{2}{c}{as \CP 1-step} & $0.13$ \\
\bottomrule
\end{tabular}
\par\vspace{0.4ex}
\parbox{\textwidth}{\scriptsize Success rate $\pm$ one standard error over the $1{,}500$ episodes. Mean of $8$: the executed chunk is the coordinate-wise mean of $8$ sampled chunks. State noise: Gaussian noise of standard deviation $0.02$, in normalized units, added to the state of the robot during training. Both: state noise and mean of $8$. The largest entry of each column is in bold. Training: with state noise. Time: one sample per chunk, with state noise. Param., Training, Time, $^{a}$, and $^{c}$: as in Table~\ref{tab:results}.}
\end{table}

\paragraph{Results on Push-T.} Table~\ref{tab:results} reports the task performance and the computational cost on Push-T, and Figure~\ref{fig:scoretime} relates the two on both tasks. On Push-T, the three proposed policies attain scores at least as high as those of the Gaussian-reference baselines \DP{}, \KRP{}, and \CP{}, with \SIP the highest overall, while using networks an order of magnitude smaller than the U-Nets of \DP and \CP and training that requires minutes rather than hours; the differences are assessed by a bootstrap over the initial conditions and the training seeds (see Appendix~\ref{app:eval} for details). The comparison of \DPR with \SIP indicates that, with $100$ or more network evaluations, the gap between \DP and the proposed policies is due mainly to the network and its training rather than to the reference.

\begin{figure}[t]
\centering
\includegraphics[width=0.41\textwidth]{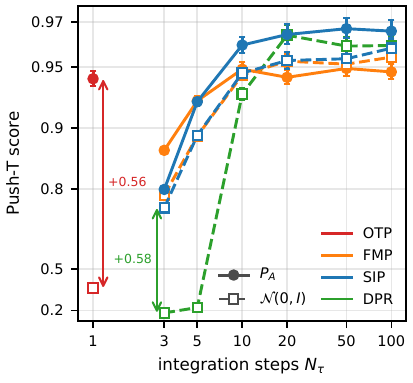}
\caption{Effect of the reference on Push-T: score against the number of integration steps $N_\tau$ with $P_A$ (filled markers, solid lines) and $\N(0, I_n)$ (open markers, dashed lines); mean $\pm$ one standard error (see Appendix~\ref{app:eval} for details). Arrows: the gain of \OTP from $\N(0, I_n)$ to $P_A$, and the gap between \DPR and \SIP with $\N(0, I_n)$ at $N_\tau = 3$.}
\label{fig:reference}
\end{figure}

\paragraph{Effect of the reference.} The reference becomes decisive when the number of network evaluations is small. The two policies that produce an action chunk with a single network evaluation, \OTP and \CP~1-step, attain comparable scores at comparable online times, but they obtain the single evaluation differently: \CP distills a diffusion teacher into a one-step student in two training stages, whereas \OTP learns the conditional OT map from $P_A$ directly, in a single stage and with a network approximately $300$ times smaller. Figure~\ref{fig:reference} isolates the effect of the reference: with $100$ integration steps, $P_A$ and $\N(0, I_n)$ attain comparable scores, whereas, as the number of steps is reduced, $P_A$ retains the score markedly better, and, for \OTP{}, replacing $P_A$ by $\N(0, I_n)$ reduces the score from $0.943$ to $0.380$. The reference $P_A$ therefore accounts for a substantial part of the performance of the proposed policies in the regime of few network evaluations, which is the regime of interest for online deployment (see Appendix~\ref{app:eval} for details).

\paragraph{Results on Square.} Square requires the nut to be placed with a precision of a few millimeters, and Table~\ref{tab:square} shows that, without modification, the single-sample \OTP falls behind \SIP and \DP{}. Its deficit was traced to two causes: the variation of a sampled chunk with the reference sample, which the mean of $8$ chunks reduces, and the extrapolation of the map away from the demonstrated states, which the state noise reduces (see Appendix~\ref{app:square} for details). The state noise improves every policy, whereas the averaging is beneficial only when the variation with the reference sample dominates the error, as for \OTP{}. With both modifications, \OTP matches the other one-step policy, \CP~1-step, in success rate and online time, and it trades a deficit of $0.03$ to $0.05$ in success rate, relative to \SIP and \DP{}, for an order of magnitude in online time.

\section{Discussion}\label{sec:discussion}

The results of Section~\ref{sec:numerics} indicate that transporting demonstrated action chunks yields policies that are competitive in task performance with \DP and \CP on both tasks, while using considerably smaller networks and shorter training. 

As for the corresponding filters in \citet{aljarrah2026generative}, \OTP is the least expensive and \SIP the most accurate of the proposed policies. In particular, \OTP produces an action chunk with a single evaluation of a network of $0.23$ million parameters on Push-T, learned in a single training stage, without the teacher and the distillation that \CP requires, which makes it suited to online settings in which an action must be produced without delay. We note that these results were obtained with deliberately small networks, of at most $8.1$ million parameters on Push-T and $24.1$ million on Square, against $65.8$ and $65.9$ million for \DP{}, respectively, drawn from a common family of fully connected networks, so that the reported scores do not represent the best attainable by the proposed policies (see Appendix~\ref{app:discussion} for details).

Moreover, the results on Square indicate that the rule by which the executed action is selected from the distribution represented by the policy deserves attention in its own right: the mean of sampled chunks benefits \OTP at an unchanged time per episode, but not \SIP or \DP, and it is not expected to be adequate for multimodal conditional distributions. Several limitations of the present study are noted: the comparison is restricted to two state-based tasks, of which Push-T has a conditional action distribution close to unimodal~\citep{mazza2026understanding}, so that the representation of multimodal conditional distributions, which motivates the formulation, is not tested directly; the baselines are used with their published settings, whereas the settings of the proposed policies were chosen for the present study; the policies are exposed to covariate shift; Proposition~\ref{prop:triangular} provides no guarantee for approximately trained maps or for the closed-loop system; and \OTP produces, at a fixed observation history, a distribution supported on at most $M$ demonstrated chunks (see Appendix~\ref{app:discussion} for the corresponding remedies). Future work will focus on these extensions, on fast and more sophisticated sampling rules that exploit the distribution of action chunks represented by the policy, on observations from images, and on the evaluation of the policies on physical systems.

\subsection*{Generative AI Use}

Generative AI tools were used for coding and experimental implementation, literature search and synthesis, and writing and editing. The authors reviewed and verified all AI-assisted content and take responsibility for the final content of the paper.

\acks{
\sloppy This work was supported by the National Science Foundation (NSF) awards EPCN-2347358 and CAREER-2440861 and the Air Force Office of Scientific Research (AFOSR) award FA9550261B219.
}

\bibliography{references}

\clearpage

\appendix

\section{Extended Related Work}\label{app:related}

In this appendix, we complete the comparison of Section~\ref{sec:intro} with the generative policies whose references or couplings are closest to those of the proposed policies; Table~\ref{tab:related} classifies them by the elements of the triangular formulation of Section~\ref{sec:triangular}.

\paragraph{Gaussian-reference policies.} The cost of the sampler of \DP (Section~\ref{sec:intro}) has motivated deterministic samplers with fewer steps~\citep{song2021denoising}, distillation into one or a few steps~\citep{prasad2024consistency,wang2025onestep}, flow-matching policies~\citep{lipman2023flow} whose number of integration steps is reduced by adaptive, consistency, mean-velocity, or minibatch optimal transport techniques~\citep{hu2024adaflow,chun2025dynamic,zhang2025flowpolicy,sheng2026mp1,sochopoulos2025fast}, and one-step generators trained by implicit maximum likelihood estimation~\citep{rana2025imle}. In all of these approaches, the generation starts from a Gaussian distribution, which carries no information about the task, so that the structure of the action distribution must be produced entirely by the learned map. In the formulation of Section~\ref{sec:triangular}, \DP and the flow-matching, consistency, and distilled policies of \citet{hu2024adaflow,black2024pi0,prasad2024consistency,wang2025onestep} take $P_Z = \N(0, I_n)$.

\paragraph{Informative references.} BRIDGeR~\citep{chen2024behavioral} connects a source policy $\pi_0(\cdot \mid s)$ to the demonstrated actions through a stochastic interpolant, and reports that an informative source improves the performance at a small number of integration steps; similar observations are made for diffusion bridges in visual navigation~\citep{ren2025prior}. On its Sphere Push-T task, the Riemannian flow-matching policy of \citet{ding2025fast} uses a uniform prior on the sphere, which, like $P_A$, is independent of both the observation and the history and outperforms a wrapped Gaussian prior at a single integration step, but which is fixed by the geometry of the action space rather than drawn from the demonstrations. A second group of methods constructs the reference from the recent behavior of the agent: WarmPrior~\citep{kang2026warmprior} centers a Gaussian at the previously executed action chunk, Streaming Flow Policy~\citep{jiang2025streaming} starts every flow at the last action, A2A~\citep{a2a2026} encodes the executed actions into a latent source, and Temporal Policy~\citep{miller2026temporal} constructs an interpolant from the state history of the robot, so that in each case the reference is paired with the target through time and its distribution at deployment is determined by the closed-loop behavior of the policy. A third group learns a reference that depends on the observation, namely a parametric Gaussian~\citep{dai2026where}, a library of Gaussians selected by a latent action~\citep{machado2026flowing}, or an embedding of the observation itself~\citep{liu2026sample,gao2026vita}. Finally, the reverse process of a pretrained \DP has been shortened by starting it from an informative chunk, namely the demonstrated action whose recorded observation is nearest to the current observation~\citep{odonchimed2025retrieve} or a chunk predicted by an auxiliary network from the current observation and the previous chunk STEP~\citep{li2026step}, and Streaming \DP~\citep{11127858}, RNR-\DP~\citep{chen2025responsive}, and real-time chunking~\citep{black2025realtime} reuse previously generated actions across re-planning instants; these methods retain the learned transport of a diffusion or flow policy trained with the Gaussian reference.

\paragraph{Optimal transport in generative policies.} The max--min formulation~\eqref{eq:maxmin} follows the approach of \citet{makkuva2020optimal} to the computation of OT maps, extended to conditional, block-triangular maps in \citet{al2023optimal,al-jarrah2024nonlinear}. OT has previously been incorporated into generative policies through minibatch couplings between a Gaussian reference and the demonstrated actions~\citep{sochopoulos2025fast}, where it was shown that a coupling that ignores the observation fails in conditional tasks, and, in reinforcement learning, through the interpretation of a diffusion policy as a transport map whose cost is given by the critic~\citep{pmlr-v267-sun25c}; in~\eqref{eq:maxmin}, by contrast, the conditional OT problem is posed between $P_A$ and $P_{A \mid S = s}$ for every $s$ simultaneously, so that the conditioning is respected by construction and no assignment problem is solved during training. We also remark that the adversarial character of~\eqref{eq:maxmin} is distinct from that of generative adversarial imitation learning~\citep{ho2016generative}, which matches the occupancy measures of states and actions through interaction with the environment, whereas~\eqref{eq:maxmin} is solved offline, from the demonstrations alone.

\paragraph{Stochastic interpolant policies.} Stochastic interpolants have previously been used for policies with a Gaussian reference~\citep{chun2025dynamic}, with the state history of the robot as reference~\citep{miller2026temporal}, and with an arbitrary source policy $\pi_0(\cdot \mid s)$~\citep{chen2024behavioral}. \SIP is the instance of the last of these constructions whose source policy is $\pi_0(\cdot \mid s) \equiv P_A$ for every $s$, up to the noise schedule, the discretization, and the network; it differs from the instances evaluated in \citet{chen2024behavioral}, whose sources are a Gaussian, a conditional variational autoencoder trained on the demonstrations, or a hand-crafted policy, in that its reference is independent of the observation and requires no model other than the data. In particular, the SDE of \citet{chen2024behavioral} is integrated by the Euler--Maruyama scheme with $5$ to $160$ steps, whereas~\eqref{eq:sifsde} is integrated by the stochastic Heun scheme with $N_\tau = 100$ steps, a choice that Section~\ref{sec:reference} revisits with $N_\tau$ between $3$ and $100$. 

\paragraph{Position of the proposed policies.} Since every policy whose reference is drawn independently of the observation, \DP included, realizes a triangular map (Appendix~\ref{sec:dp}), the proposed policies are distinguished not by the triangular structure but by the reference $P_A$, which requires no auxiliary model, no domain knowledge, and no pairing of the training data, and whose distribution at deployment coincides with its distribution during training; moreover, \OTP combines this reference with a single network evaluation, learned in a single training stage as a conditional OT map, whereas existing one-step policies start from a Gaussian. We note that \SIP is formally the instance of BRIDGeR~\citep{chen2024behavioral} whose source policy is the action marginal, up to the noise schedule, the discretization, and the network, so that its contribution lies in this choice of reference rather than in the interpolant itself.

\begin{table}[t]
\caption{Generative policies related to the proposed policies, classified by the elements of the triangular formulation of Section~\ref{sec:triangular}; the entries are taken from the cited works, and these policies are not evaluated in this paper.}
\label{tab:related}
\centering
\footnotesize
\setlength{\tabcolsep}{3pt}
\begin{tabular}{@{}>{\raggedright\arraybackslash}p{0.19\textwidth}
  >{\raggedright\arraybackslash}p{0.13\textwidth}
  >{\raggedright\arraybackslash}p{0.13\textwidth}
  >{\raggedright\arraybackslash}p{0.13\textwidth}
  >{\raggedright\arraybackslash}p{0.13\textwidth}
  c
  >{\raggedright\arraybackslash}p{0.13\textwidth}@{}}
\toprule
Policy & Reference $P_Z$ & Coupling during training & Learning & Inference & NFE & Auxiliary model \\
\midrule
IMLE Policy~\citep{rana2025imle} & Gaussian latent & nearest generated sample & implicit maximum likelihood & one-shot map & $1$ & none \\
COT Policy~\citep{sochopoulos2025fast} & Gaussian & conditional minibatch OT & flow matching & ODE & $2$ & OT solver in training \\
DA-SIP~\citep{chun2025dynamic} & Gaussian & independent & interpolant regression & ODE or SDE$^\dagger$ & $1$--$100$ & difficulty classifier \\
BRIDGeR~\citep{chen2024behavioral} & source policy $\pi_0(\cdot \mid s)$: Gaussian, CVAE, or heuristic & conditionally independent given $s$ & interpolant regression & SDE$^\dagger$, Euler--Maruyama & $5$--$160^{\ddagger}$ & source policy \\
WarmPrior~\citep{kang2026warmprior} & Gaussian centered at the previous chunk & paired in time & flow matching & ODE & $1$--$9$ & none \\
Temporal Policy~\citep{miller2026temporal} & state history & paired in time & interpolant regression & ODE or SDE$^\dagger$ & $10$ & none \\
LeaP~\citep{dai2026where} & learned Gaussian, conditioned on proprioception & conditionally independent given $s$ & flow matching, likelihood and contrastive alignment & ODE & $6^{\ddagger}$ & prior network \\
BridgePolicy~\citep{liu2026sample} & embedding of the observation & paired & diffusion bridge & bridge sampler$^\dagger$ & $10$ & fusion module, semantic aligner \\
\bottomrule
\end{tabular}
\par\vspace{0.4ex}
\parbox{\textwidth}{\scriptsize NFE: network evaluations per action chunk, as stated by each work; DA-SIP and BRIDGeR report the number of integration steps instead. $^\dagger$Stochastic inference: repeated evaluation at a fixed reference sample yields different action chunks. $^{\ddagger}$In addition to one evaluation of the source policy. CVAE: conditional variational autoencoder. 
}
\end{table}

\section{Proof of Proposition~\ref{prop:triangular}}\label{app:proof}

\begin{proof}
\textbf{Step 1:} Let $(S, Z) \sim P_S \otimes P_Z$, and let $\varphi : \R^{m} \to \R$ and $\psi : \R^{n} \to \R$ be bounded and measurable. The hypothesis $G_{\#}(P_S \otimes P_Z) = P_{S,A}$ gives
$\Expect[\varphi(S)\, \psi(T(S, Z))] = \Expect_{P_{S,A}}[\varphi(S)\, \psi(A)]$.

\textbf{Step 2:} Since $S$ and $Z$ are independent, the left-hand side equals the first of the integrals below, while the disintegration of $P_{S,A}$ with respect to $P_S$ shows that the right-hand side equals the second:
\begin{equation*}
  \int \varphi(s) \Bigl(\int \psi \, \ud\bigl[T(s,\cdot)_{\#}P_Z\bigr]\Bigr) P_S(\ud s)
  \quad\text{and}\quad
  \int \varphi(s) \Bigl(\int \psi\, \ud P_{A\mid S=s}\Bigr) P_S(\ud s).
\end{equation*}

\textbf{Step 3:} Since $\varphi$ is arbitrary, the two inner integrals coincide for $P_S$-almost every $s$. Applying this conclusion to each element $\psi$ of a countable convergence-determining class of bounded continuous functions, and using the fact that a countable union of $P_S$-null sets is $P_S$-null, one verifies that $T(s,\cdot)_{\#}P_Z$ and $P_{A\mid S=s}$ coincide for $P_S$-almost every $s$, which concludes the proof.
\end{proof}

\section{Derivation of the Proposed Policies}\label{app:derivation}

In this appendix, we derive the three proposed policies summarized in Section~\ref{sec:proposed}.

\subsection{\OTPLong{} (\OTP{})}\label{sec:otf}

\OTP{}, derived from the optimal transport filter of \citet{al-jarrah2024nonlinear}, selects $T(s, \cdot)$ to be the OT map from $P_A$ to $P_{A \mid S = s}$ for every observation history $s$. If the OT cost is quadratic, the OT map is obtained as the solution of the max--min problem~\citep{al2023optimal,al-jarrah2024nonlinear}
\begin{equation}\label{eq:maxmin}
 \max_{f}\,\min_{T}\;\; \mathcal{J}(f, T)
 := \Expect\bigl[ f(S, A)\bigr]
 + \Expect\Bigl[ \tfrac{1}{2}\bigl\| T(S, \bar A) - \bar A \bigr\|^2
 - f\bigl(S, T(S, \bar A)\bigr) \Bigr],
\end{equation}
where the first expectation is taken with respect to $P_{S,A}$, the second with respect to the independent coupling $(S, \bar A) \sim P_S \otimes P_A$, and $f : \R^{m}\times\R^{n} \to \R$ is a potential function. In a numerical implementation, $f$ and $T$ are parameterized by residual networks and trained adversarially (Appendix~\ref{app:impl}); at deployment, the policy~\eqref{eq:policy} requires a single evaluation of $T$ per action chunk, and $f$ is discarded.

\subsection{\FMPLong{} (\FMP{})}\label{sec:fmf}

In \FMP{}, derived from the flow matching filter of \citet{aljarrah2026generative}, the map $T$ is realized as the time-one flow of an ODE~\citep{lipman2023flow}: for each $(s, \bar a) \in \R^{m}\times\R^{n}$, we set $T(s, \bar a) = \xi_1$, where
\begin{equation}\label{eq:fmfode}
 \frac{\ud \xi_\tau}{\ud \tau} = b_\theta\bigl(s, \xi_\tau, \tau\bigr), \qquad \xi_0 = \bar a,
\end{equation}
and $b_\theta : \R^{m}\times\R^{n}\times[0, 1] \to \R^{n}$ is a learned vector field. The observation history $s$ enters~\eqref{eq:fmfode} as a parameter and is held fixed along the flow, so that the associated map is triangular in the sense of~\eqref{eq:triangular-map} by construction. The vector field is obtained as the solution to the regression problem
\begin{equation}\label{eq:fmfloss}
 \min_{\theta}\; \Expect\Bigl[\bigl\| b_\theta(S, A_\tau, \tau) - \partial_\tau A_\tau \bigr\|^2\Bigr],
\end{equation}
where the expectation is taken over $\tau \sim \mathrm{Unif}[0, 1]$, $(S, A) \sim P_{S,A}$, and $\bar A \sim P_A$ drawn independently of $(S, A)$, and where
\begin{equation}\label{eq:lininterp}
 A_\tau = (1-\tau)\,\bar A + \tau A, \qquad \tau \in [0, 1],
\end{equation}
is the linear interpolant between the reference and the target, whose derivative $\partial_\tau A_\tau = A - \bar A$ is available in closed form. The policy~\eqref{eq:policy} is evaluated by integrating~\eqref{eq:fmfode} from $\xi_0 = \bar A_k$ with $s = S_k$ fixed, by the Heun scheme with $N_\tau = 100$ uniform steps, that is, with $200$ network evaluations.

\subsection{\SIPLong{} (\SIP{})}\label{sec:sif}

\SIP{}, derived from the stochastic interpolant filter of \citet{aljarrah2026generative}, generalizes \FMP along two independent axes~\citep{albergo2025}: the interpolant connecting the reference to the target may carry noise, and the generation may be carried out by an SDE rather than an ODE. The linear interpolant~\eqref{eq:lininterp} is generalized to
\begin{equation}\label{eq:sifinterp}
 A_\tau = (1-\tau)\,\bar A + \tau A + \gamma(\tau)\, \zeta, \qquad
 \gamma(\tau) = \varepsilon\sqrt{\tau(1-\tau)},
\end{equation}
where $\zeta \sim \N(0, I_n)$ is independent of $(S, A, \bar A)$ and $\varepsilon \ge 0$ sets the noise level; the case $\varepsilon = 0$ recovers~\eqref{eq:lininterp}. The velocity field is learned exactly as in~\eqref{eq:fmfloss}, with the regression target now $\partial_\tau A_\tau = A - \bar A + \dot\gamma(\tau)\, \zeta$. A second function $\eta_\theta$, realized as a second output head of the same network, is trained to recover the latent noise,
\begin{equation}\label{eq:sifden}
 \min_{\theta}\; \Expect\Bigl[\bigl\| \eta_\theta(S, A_\tau, \tau) - \zeta \bigr\|^2\Bigr],
\end{equation}
whose minimizer $\eta$ determines the conditional score of the interpolant~\citep{albergo2025}, namely $\nabla_a \log \rho_\tau(a \mid s) = -\eta(s, a, \tau)/\gamma(\tau)$ whenever $\gamma(\tau) > 0$, where $\rho_\tau(\cdot \mid s)$ denotes the density of $A_\tau$ given $S = s$. Given the score, the deterministic flow~\eqref{eq:fmfode} may be replaced by the SDE
\begin{equation}\label{eq:sifsde}
 \ud \xi_\tau = \bar b_\theta\bigl(s, \xi_\tau, \tau\bigr)\, \ud\tau + \lambda\, \gamma(\tau)\, \ud W_\tau, \qquad \xi_0 = \bar a,
\end{equation}
with drift $\bar b_\theta(s, a, \tau) = b_\theta(s, a, \tau) - \tfrac{1}{2}\lambda^2 \gamma(\tau)\, \eta_\theta(s, a, \tau)$, where $W_\tau$ denotes a standard Wiener process on $\R^{n}$ and $\lambda \ge 0$ is a parameter that controls the diffusion coefficient. The score correction in~\eqref{eq:sifsde} compensates exactly for the injected noise, so that, at the minimizers of~\eqref{eq:fmfloss} and~\eqref{eq:sifden}, \eqref{eq:sifsde} and~\eqref{eq:fmfode} share the time-$\tau$ marginals of the interpolant~\eqref{eq:sifinterp} conditioned on $S = s$ for every choice of $\lambda$; in particular, $\xi_1 \sim P_{A \mid S = s}$, since the interpolant starts from $P_A$ and ends at $P_{A \mid S = s}$ by the independence of $\bar A$ and $S$.

In the experiments, \SIP is realized by the SDE~\eqref{eq:sifsde} with $\varepsilon > 0$ and $\lambda = 1$, integrated by a stochastic Heun scheme with $N_\tau = 100$ uniform steps, so that \SIP, like \FMP, requires $200$ network evaluations per action chunk (Appendix~\ref{app:impl}). We note that \SIP is formally the instance of BRIDGeR~\citep{chen2024behavioral} whose source policy is the action marginal (Appendix~\ref{app:related}).

\section{Gaussian-Reference Policies}\label{app:baselines}

In this appendix, we describe the three policies with which the proposed policies are compared in Section~\ref{sec:numerics}, namely \KRP and the two baselines \DP and \CP, together with \DPR, a variant of \DP realized with the network and the sampler of \FMP and \SIP, all of which transport the Gaussian reference $P_Z = \N(0, I_n)$, and identify them as instances of the triangular formulation of Section~\ref{sec:triangular}. Table~\ref{tab:methods} summarizes the seven policies evaluated in this paper.

\subsection{\KRPLong{} (\KRP{})}\label{sec:krf}

As an instance with the Gaussian reference $P_Z = \N(0, I_n)$, which is also the reference of \DP and \CP (Appendices~\ref{sec:dp} and~\ref{sec:cp}), we consider \KRP{}, derived from the stochastic map filter of \citet{spantini2022coupling}, which is based on the Knothe--Rosenblatt rearrangement and is considered in \citet{aljarrah2026generative}, in which the triangular map is learned in the inverse direction, from the data to the reference, by maximum likelihood. More precisely, \KRP learns a map $\Phi_\theta : \R^{m}\times\R^{n} \to \R^{n}$ that is lower triangular and monotone in the action chunk, with components
\begin{equation}\label{eq:krfmap}
  \Phi_{\theta, j}(s, a) = c_j(s, a_{<j}) + \int_0^{a_j} \Bigl( \mathrm{softplus}\bigl(g_j(s, a_{<j}, t)\bigr) + \delta \Bigr)\, \ud t, \qquad j = 1, \ldots, n,
\end{equation}
where $a_{<j} = (a_1, \ldots, a_{j-1})$, $c_j$ and $g_j$ are neural networks, and $\delta > 0$, so that $\Phi_{\theta, j}$ is strictly increasing in $a_j$ by construction~\citep{spantini2022coupling,baptista2024representation}. The parameters are obtained by maximizing the likelihood of the training data under the push-forward of the reference, that is, by solving
\begin{equation}\label{eq:krfloss}
  \min_{\theta}\; \Expect\Bigl[ \tfrac{1}{2}\bigl\|\Phi_\theta(S, A)\bigr\|^2 - \sum_{j=1}^{n} \log \partial_{a_j} \Phi_{\theta, j}(S, A) \Bigr],
\end{equation}
where $(S, A) \sim P_{S,A}$ and the partial derivatives are the integrands of~\eqref{eq:krfmap}, available in closed form. At the minimizer, the map $(s, a) \mapsto (s, \Phi_\theta(s, a))$ transports $P_{S,A}$ to $P_S \otimes \N(0, I_n)$, so that its inverse is triangular in the sense of~\eqref{eq:triangular-map}, with $T(s, \cdot) = \Phi_\theta(s, \cdot)^{-1}$, and Proposition~\ref{prop:triangular} applies; since $\Phi_\theta(s, \cdot)$ is moreover lower triangular and monotone, $T(s, \cdot)$ is the Knothe--Rosenblatt rearrangement of $\N(0, I_n)$ to $P_{A \mid S = s}$~\citep{marzouk2017sampling}. The policy is therefore $\hat A_k = \Phi_\theta(S_k, \cdot)^{-1}(Z_k)$, where, as in the analysis step of \citet{spantini2022coupling}, the reference sample is $Z_k = \Phi_\theta(S^i, A^i)$ for a demonstrated pair $(S^i, A^i)$ drawn uniformly from the training data, which is distributed as $\N(0, I_n)$ at the minimizer of~\eqref{eq:krfloss}, and the inverse is computed coordinate by coordinate, each coordinate being the root of a scalar monotone equation, by a safeguarded Newton iteration (Appendix~\ref{app:impl}). \KRP differs from the analysis step of \citet{spantini2022coupling} in that the functions $c_j$ and $g_j$ are neural networks rather than linear combinations of basis functions, so that~\eqref{eq:krfloss} is not convex. On Push-T, drawing $Z_k$ from $\N(0, I_n)$ instead changes the score of \KRP from $0.914$ to $0.918$, a difference of $0.004 \pm 0.005$. In contrast with \DP, \KRP is a normalizing flow~\citep{marzouk2017sampling} rather than a denoising model, which provides a second Gaussian-reference comparison. The Knothe--Rosenblatt rearrangement filter is compared in \citet{aljarrah2026generative} as an established alternative to the filters from which the proposed policies are derived.

\subsection{Diffusion Policy (\DP) as a Gaussian-reference instance}\label{sec:dp}

\DP~\citep{chi2023diffusion} is the instance of the formulation of Section~\ref{sec:triangular} with $P_Z = \N(0, I_n)$. Given a noise schedule $0 < \beta_1 < \cdots < \beta_L < 1$ and $\bar\alpha_\ell := \prod_{j \le \ell}(1 - \beta_j)$, a network $e_\theta$ is trained to recover the noise from the noised action chunk,
\begin{equation}\label{eq:ddpmloss}
  \min_{\theta}\;
  \Expect\Bigl[ \bigl\| e_\theta\bigl(\sqrt{\bar\alpha_\ell}\, A + \sqrt{1 - \bar\alpha_\ell}\, Z,\; \ell,\; S\bigr) - Z \bigr\|^2 \Bigr],
\end{equation}
where $\ell \sim \mathrm{Unif}\{1, \ldots, L\}$, $(S, A) \sim P_{S,A}$, and $Z \sim \N(0, I_n)$ are mutually independent, and an action chunk is generated from $A^{(L)} \sim \N(0, I_n)$ by $L = 100$ steps of the ancestral sampler (Appendix~\ref{app:impl}), a stochastic map that leaves $s$ unchanged and is therefore triangular. The difference from the proposed policies lies not in the conditioning structure but in the reference, in a path prescribed by a noising process that is independent of the data, rather than by an OT map or by an interpolant between demonstrated actions, and in the $L$ evaluations of $e_\theta$, a one-dimensional convolutional U-Net; we remark that~\eqref{eq:sifinterp} with a Gaussian sample in place of $\bar A$ defines the Gaussian-reference counterpart of \SIP~\citep{albergo2025}, which isolates the effect of the reference (Section~\ref{sec:reference}).

\subsection{Diffusion Policy with a Residual Network (\DPR)}\label{sec:dpr}

Since \DP differs from the proposed policies not only in its reference but also in its network, its sampler, and its settings, we consider, in addition, a variant of \DP, which we refer to as \DPR, that retains the defining elements of \DP, namely the Gaussian reference $\N(0, I_n)$ drawn independently of the observation history, the cosine noise schedule~\citep{nichol2021improved}, and the noise-prediction objective~\eqref{eq:ddpmloss}, and takes every other element from \FMP and \SIP, namely the residual network, the training procedure, and the integration in a continuous artificial time. More precisely, with $\tau \in [0, 1]$ running from the reference ($\tau = 0$) to the data ($\tau = 1$), the noised action chunk is $A_\tau = \alpha(\tau)\, A + \sigma(\tau)\, Z$, where $Z \sim \N(0, I_n)$ and $\alpha(\tau)^2 + \sigma(\tau)^2 = 1$ are given by the cosine schedule, and a network $e_\theta(S, A_\tau, \tau)$ is trained to recover $Z$, as in~\eqref{eq:ddpmloss} with the discrete index $\ell$ replaced by the continuous time $\tau \sim \mathrm{Unif}[0, 1]$. An action chunk is generated from $Z_k \sim \N(0, I_n)$ by integrating the corresponding probability-flow ODE, in the coordinates of the deterministic sampler of \citet{song2021denoising}, by the Heun scheme of \citet{karras2022elucidating} with $N_\tau = 100$ uniform steps, that is, with $200$ network evaluations per action chunk; as in \DP, the predicted chunk is clipped to the normalized range of the demonstrated actions at every evaluation. The resulting map leaves $s$ unchanged and is therefore triangular in the sense of~\eqref{eq:triangular-map}. On Push-T, \DPR evaluates a network of $8.5$ million parameters (width $1024$, four residual blocks) with the horizons $(H_o, H_p, H_a) = (1, 8, 8)$, and, on Square, a network of $18.2$ million parameters (width $1728$, three residual blocks) with the horizons $(2, 4, 4)$. The network uses layer normalization, the mish activation, and a time embedding of dimension $10$, and it is trained with AdamW and a cosine schedule, without an exponential moving average of the weights; on Push-T, at the learning rate $3.68 \times 10^{-4}$ with $2\%$ warm-up, a weight decay of $0.1$, gradient clipping at $0.1$, a batch size of $256$, and $65{,}536$ iterations, and, on Square, at the learning rate $4.24 \times 10^{-4}$ with $5\%$ warm-up, a weight decay of $0.01$, gradient clipping at $0.5$, a batch size of $1088$, and $262{,}144$ iterations.

On Push-T, under the protocol of Section~\ref{sec:pusht}, \DPR attains a score of $0.963 \pm 0.004$, with a seed spread of $0.009$, $56.6\%$ successes, and $3.1\%$ failures; its paired difference from \SIP is $-0.002 \pm 0.005$. It is trained in $2.4$ minutes per seed and requires $2.84$ seconds per episode. On Square, without modification, it succeeds in $89.6\% \pm 0.8\%$ of the episodes, and in $89.1\% \pm 0.8\%$ with the mean of $8$ chunks. With state noise, it succeeds in $94.2\% \pm 0.6\%$ of the episodes, and in $93.5\% \pm 0.6\%$ with the mean of $8$ chunks, which is comparable with \SIP and \DP under the same training; it is trained in $44$ minutes per seed without state noise and in $46$ minutes with it, and requires $1.05$ seconds per episode with state noise, against $1.24$ for \SIP and \DP (Table~\ref{tab:square}). On Push-T, its score is retained down to $10$ integration steps ($0.932$) and drops to $0.226$ with $5$ steps (Figure~\ref{fig:reference}).

\subsection{Consistency Policy (\CP)}\label{sec:cp}

\CP~\citep{prasad2024consistency} is likewise an instance of the formulation of Section~\ref{sec:triangular} with $P_Z = \N(0, I_n)$, obtained in two training stages: a diffusion model in the parameterization of \citet{karras2022elucidating} is first trained as a \emph{teacher}, and a \emph{student} network, initialized from the teacher, is then distilled by the consistency trajectory objective of \citet{prasad2024consistency}, so that it maps a noised action chunk at a given noise level directly to a chunk at a lower noise level along the trajectories of the probability-flow ODE of the teacher. At deployment, \CP~1-step maps a Gaussian sample at the largest noise level to an action chunk in a single evaluation of the student, and \CP~3-step re-noises this output twice, to intermediate noise levels, and maps it back each time, which yields a stochastic sampler with three evaluations. Both samplers leave $s$ unchanged and are therefore triangular in the sense of~\eqref{eq:triangular-map}, the reference of \CP~3-step being extended by the re-noising variables. In contrast with \OTP, the other one-step policy of this paper, \CP starts from a Gaussian, requires a teacher and a second training stage, and evaluates a U-Net of $69.5$ million parameters (Table~\ref{tab:cost}); the details of the implementation are given in Appendix~\ref{app:impl}.

\section{Implementation Details}\label{app:impl}

In this appendix, we describe the implementation of the six policies and of the deployment protocol; the hyperparameters of the proposed policies are listed in Table~\ref{tab:hyper} of Appendix~\ref{app:tuning}.

\paragraph{Training data.} The training data of Section~\ref{sec:problem} are obtained by collecting the pairs~\eqref{eq:chunks} over every time instant of every episode, with the first observation and the last action of an episode repeated where the windows extend beyond its boundaries; at deployment, the observation history is likewise completed by repeating the first observation of the episode. Since the observation histories of the data set are, in general, distinct, the empirical conditional distribution at each of them is a Dirac measure, so that the stochasticity of a learned policy relies on the generalization of the map across neighboring observation histories.

\paragraph{Proposed policies.} In \OTP, the potential $f$ and the map $T$ are residual networks whose input is the concatenation of their two arguments, and the map is parameterized as the displacement $T(s, \bar a) = \bar a + \Delta_\theta(s, \bar a)$, with the output layer of $\Delta_\theta$ initialized at zero; at every iteration, a minibatch of training pairs is drawn with replacement, the observation histories are permuted within the minibatch to realize the independent coupling, and $K_{\mathrm{in}} = 25$ gradient steps on $T$ precede one gradient step on $f$. In \FMP and \SIP, the networks are residual multilayer perceptrons with layer normalization, whose input is the concatenation of $a$, a sinusoidal embedding of $\tau$, and $s$; the network of \SIP has the two output heads $b_\theta$ and $\eta_\theta$, and that of \FMP the first only. Every step of the Heun scheme of \FMP and of the stochastic Heun scheme of \SIP evaluates the network twice, and, at every training iteration, the index of the target pair and the index of the reference chunk are drawn independently, with replacement, from the training data; the two terms~\eqref{eq:fmfloss} and~\eqref{eq:sifden} of \SIP are weighted equally. In~\eqref{eq:fmfode}, the artificial time $\tau \in [0, 1]$ is distinct from the time $k$ of the system. In~\eqref{eq:sifsde}, setting $\lambda = 0$ recovers the ODE. The equivalence of~\eqref{eq:sifsde} and~\eqref{eq:fmfode} holds for the exact fields; with the learned fields $b_\theta$ and $\eta_\theta$ and a numerical integrator, the two samplers agree only up to the approximation and discretization errors.

\paragraph{Training.} The hyperparameters of the proposed policies are listed in Table~\ref{tab:hyper} (Appendix~\ref{app:tuning}). The baselines are used with their published settings: \DP with the configuration published in \citet{chi2023diffusion}, trained for $505{,}000$ gradient steps ($5{,}000$ epochs at a batch size of $256$) and deployed with the exponential moving average of its weights, and \CP~\citep{prasad2024consistency} with the settings shared by its released robomimic configurations and the U-Net, the horizons, and the batch size of \DP, its teacher~\citep{karras2022elucidating} and its student each being trained for $150{,}000$ gradient steps (see the paragraph on \CP below); the student is deployed with one step (\CP~1-step) or, re-noising its output twice, with three steps (\CP~3-step). All six policies are trained on all $206$ demonstrations (Appendix~\ref{app:eval}), and the horizons $(H_o, H_p, H_a)$ are $(1, 4, 4)$ for \OTP and \FMP, $(1, 8, 8)$ for \SIP and \KRP, and $(2, 16, 8)$ for \DP and \CP.

\paragraph{\DP.} The ancestral sampler of Appendix~\ref{sec:dp} applies, for $\ell = L, \ldots, 1$, the step
\begin{equation}\label{eq:ddpmstep}
  A^{(\ell-1)} = \tfrac{1}{\sqrt{1-\beta_\ell}}\Bigl( A^{(\ell)} - \tfrac{\beta_\ell}{\sqrt{1-\bar\alpha_\ell}}\, e_\theta\bigl(A^{(\ell)}, \ell, s\bigr) \Bigr) + \sigma_\ell\, \zeta_\ell,
\end{equation}
with $\zeta_\ell \sim \N(0, I_n)$ and $\sigma_\ell^2 = \beta_\ell (1 - \bar\alpha_{\ell-1})/(1 - \bar\alpha_\ell)$, the output being $A^{(0)}$. In the implementation of \citet{chi2023diffusion}, used here, every step first forms the estimate of the clean chunk implied by $e_\theta$ and clips it to the normalized range $[-1, 1]^n$, so that the step coincides with~\eqref{eq:ddpmstep} whenever the clipping is inactive. The noise schedule is the squared-cosine schedule of \citet{nichol2021improved}, with $\beta_\ell$ capped at $0.999$; the U-Net, conditioned on $s$ and $\ell$ by feature-wise linear modulation~\citep{perez2018film}, has the channel widths $256$, $512$, and $1024$, a kernel size of $5$, and diffusion-step embeddings of dimension $256$, and it is trained with AdamW at the learning rate $10^{-4}$, with $500$ warm-up steps and a weight decay of $10^{-6}$, and an exponential moving average of the weights is maintained for deployment, all as in the published configuration.

\paragraph{\CP.} The implementation is a self-contained module extracted from the code of \citet{prasad2024consistency}, and it was verified against that code, to numerical precision, for the network, the noise scalings, the losses, the initialization of the student from the teacher, and the sampler. The teacher is a diffusion model in the parameterization of \citet{karras2022elucidating}, with noise levels in $[0.02, 80]$ and a pseudo-Huber loss; the student, initialized from the teacher, is distilled by the consistency trajectory objective of \citet{prasad2024consistency}, combined in equal proportions with a denoising score-matching term. \CP~1-step maps the noise level $80$ to the level $0.02$ in one evaluation, and \CP~3-step re-noises the output to the levels $27$ and $54$ and maps it back each time. Each stage is trained for $150{,}000$ gradient steps at a batch size of $256$, as for \DP{}, so that the two stages together require approximately the training time of \DP.

\paragraph{\KRP.} The $n$ components of~\eqref{eq:krfmap} are evaluated in parallel: the networks $c_j$ and $g_j$ of all components are stacked into batched weight tensors, and the triangular structure is imposed by a fixed binary mask on the input $(s, a)$, so that component $j$ receives $s$ and $a_{<j}$ only. Every $c_j$ and $g_j$ is a multilayer perceptron with three hidden layers of width $320$ and the tanh activation, $\delta = 10^{-3}$, and the integral of~\eqref{eq:krfmap} is evaluated by the Gauss--Legendre rule with $16$ nodes. The map is trained on~\eqref{eq:krfloss} with AdamW at the learning rate $2.88 \times 10^{-3}$ and a weight decay of $10^{-3}$, a cosine schedule with $5\%$ warm-up, gradient clipping at $5.0$, a batch size of $320$, and $32{,}768$ iterations, with the horizons $(1, 8, 8)$, so that $n = 16$. At deployment, coordinate $j$ of $\Phi_\theta(S_k, \cdot)^{-1}(Z_k)$ is the root of the scalar equation $\Phi_{\theta, j}(S_k, \hat a_{<j}, a_j) = Z_{k, j}$, which is bracketed in $[-3, 3]$, in normalized coordinates, with the bracket doubled until it contains the root, and solved by at most $10$ safeguarded Newton steps, in which a step that leaves the bracket is replaced by a bisection step; the derivative required by Newton's method is the integrand of~\eqref{eq:krfmap}. The reference sample is $\Phi_\theta(S^i, A^i)$ for a demonstrated pair drawn uniformly. On Square, with $n = 40$, every $c_j$ and $g_j$ has three hidden layers of width $320$ with residual connections and the mish activation, and the map is trained with AdamW at the learning rate $3 \times 10^{-3}$ and a weight decay of $10^{-2}$, a cosine schedule without warm-up, gradient clipping at $1.0$, a batch size of $576$, and $49{,}152$ iterations, with an exponential moving average of the weights of decay $0.999$ and the horizons $(2, 4, 4)$.

\begin{algorithm2e}[htbp]
\caption{Deployment of a generative transport policy}
\label{alg:deploy}
\DontPrintSemicolon\SetCommentSty{textit}
\KwIn{map $T$, learned on normalized coordinates; demonstrated action chunks $\{A^i\}_{i=1}^{M}$; horizons $H_o$, $H_p$, and $H_a$; episode length $K_{\max}$}
Observe $o_0$, set $o_{-H_o+1} = \cdots = o_{-1} := o_0$, and set $k \gets 0$\;
\While{$k < K_{\max}$ and the task is not completed}{
  $S_k \gets (o_{k-H_o+1}, \ldots, o_k)$; draw $i \sim \mathrm{Unif}\{1, \ldots, M\}$ and set $\bar A_k \gets A^i$\;
  $(\hat a_k, \ldots, \hat a_{k+H_p-1}) \gets T(S_k, \bar A_k)$ \tcp*{normalized inputs, denormalized output}
  \For{$j = k, \ldots, k + H_a - 1$}{
    Apply $\mathrm{clip}(\hat a_j)$, clipped per coordinate to the demonstrated range, and observe $o_{j+1}$\;
  }
  $k \gets k + H_a$\;
}
\end{algorithm2e}

\paragraph{Deployment.} Algorithm~\ref{alg:deploy} summarizes the deployment protocol of Section~\ref{sec:deployment}. The observation histories and the action chunks are normalized affinely to $[-1, 1]$ per coordinate, using the ranges of the demonstrations, so that the quadratic cost of~\eqref{eq:maxmin} and the isotropic noise of~\eqref{eq:sifinterp} are defined in these coordinates, and the executed actions are clipped, per coordinate, to the range of the demonstrated actions. Since the observation history is generated in closed loop, the protocol of Section~\ref{sec:deployment} reproduces the independence of the reference from the observation history and the distribution $P_A$ of the reference, but the coupling $P_S \otimes P_A$ itself only insofar as $S_k$ remains distributed as in the demonstrations. The alternative in which the reference is the action chunk produced at the previous re-planning instant violates this independence, since that chunk is correlated with the current observation history, and it was markedly inferior in preliminary experiments. The history-based policies of \citet{kang2026warmprior,miller2026temporal,jiang2025streaming} avoid the mismatch of the previous-chunk reference by training on references paired with their targets through time, although the distribution of the reference at deployment is then determined by the closed-loop behavior of the policy. The clipping box of Algorithm~\ref{alg:deploy}, namely the per-coordinate range of the $25{,}650$ demonstrated actions, is $[12, 511] \times [25, 511]$ in pixels, which covers almost the whole workspace $[0, 512]^2$; the clipping therefore confines each coordinate of the action to the demonstrated range, rather than the action to a demonstrated region. 

\paragraph{Conditioning on the observations only.} We remark that the conditioning variable $S_k$ may, in principle, also contain the most recently executed actions $a_{k-H_o}, \ldots, a_{k-1}$; this is, however, not recommended, since a policy conditioned on its own past actions tends to copy them rather than respond to the observations, a failure mode known as the copycat problem~\citep{wen2020fighting} and, more generally, as causal confusion~\citep{dehaan2019causal}. Accordingly, we take $S_k$ to consist of observations only. Finally, the policies are conditioned on the observations only (Section~\ref{sec:problem}), since conditioning on the most recently executed actions degraded the performance in preliminary experiments, in agreement with the causal confusion of \citet{dehaan2019causal} and the copycat problem of \citet{wen2020fighting}.

\section{Hyperparameters}\label{app:tuning}

Table~\ref{tab:hyper} lists the hyperparameters of the proposed policies on Push-T. These values were selected with the Bayesian optimization package SMAC3~\citep{lindauer2022smac3}, which searches over the network and training hyperparameters for the configuration that maximizes the evaluation score; the same procedure yields the values on Square in Table~\ref{tab:hypersquare}.

\begin{table}[t]
\caption{Hyperparameters of the proposed policies on Push-T.}

\label{tab:hyper}
\centering
\footnotesize
\setlength{\tabcolsep}{4pt}
\begin{tabular}{@{}>{\raggedright\arraybackslash}p{0.22\textwidth}
  >{\raggedright\arraybackslash}p{0.27\textwidth}
  >{\raggedright\arraybackslash}p{0.22\textwidth}
  >{\raggedright\arraybackslash}p{0.22\textwidth}@{}}
\toprule
 & \OTP & \FMP & \SIP \\
\midrule
Horizons $(H_o, H_p, H_a)$ & $(1, 4, 4)$ & $(1, 4, 4)$ & $(1, 8, 8)$ \\
Networks & $f$: width $576$, $3$ residual blocks; $T$: width $192$, $3$ residual blocks & width $896$, $5$ residual blocks, layer normalization & width $512$, $5$ residual blocks, layer normalization \\
Activation & mish & mish & relu \\
Time embedding (dimension) & -- & $20$ & $10$ \\
Optimizer & AdamW, weight decay $0$ & AdamW, weight decay $0.2$ & AdamW, weight decay $0.01$ \\
Learning rate & $5.97 \times 10^{-5}$ ($f$), $7.23 \times 10^{-4}$ ($T$) & $9.33 \times 10^{-4}$ & $1.92 \times 10^{-3}$ \\
Schedule & cosine, $5\%$ warm-up & cosine, $2\%$ warm-up & cosine, $2\%$ warm-up \\
Gradient clipping & $0.25$ & $1.0$ & $0.25$ \\
Batch size & $576$ & $256$ & $320$ \\
Iterations & $16{,}384$, with $K_{\mathrm{in}} = 25$ & $16{,}384$ & $65{,}536$ \\
Noise level $\varepsilon$ & -- & $0$ & $0.115$ \\
Sampler & one evaluation & Heun, $N_\tau = 100$ & stochastic Heun, $N_\tau = 100$, $\lambda = 1$ \\
\bottomrule
\end{tabular}
\end{table}

\section{Additional Evaluation Details}\label{app:eval}

\paragraph{Environment and computation.} On Push-T, the action $a_k \in \R^{2}$ is the commanded position of the end effector, applied at $10$\,Hz, and the observation $o_k \in \R^{20}$ comprises the planar positions of nine keypoints of the block and of the end effector; the environment is the reference implementation of \citet{chi2023diffusion}. Two modifications of the code of \citet{chi2023diffusion} were required for compatibility with current software versions, neither of which affects the simulation: the \texttt{gym} import is replaced by \texttt{gymnasium}, and a deprecated \texttt{matplotlib} colormap accessor is replaced by its current equivalent. The experiments were performed on a workstation with three NVIDIA RTX 5000 Ada Generation GPUs and an AMD Ryzen Threadripper PRO 7965WX CPU with $24$ cores, with PyTorch~2.11, Gymnasium~1.3.0, and pymunk~6.11.1. The training times of Tables~\ref{tab:results} and~\ref{tab:square} were measured with one training run per GPU and vary by less than $8\%$ across training seeds; the split of the training time of \CP between the teacher and the student is estimated from the times at which the checkpoints of the two stages were written. In the evaluation, the $500$ episodes of each training seed are distributed over the three GPUs, so that the time per episode of Tables~\ref{tab:results} and~\ref{tab:square} is a measure of throughput, which also depends on the number of steps per episode, smaller for a policy that completes the task earlier.

\paragraph{Figure~\ref{fig:overview}.} Figure~\ref{fig:overview} illustrates the policy~\eqref{eq:policy} for \OTP on Push-T: the reference chunks, spread over the workspace, are transported to chunks near the end effector, which depend on the observation history. The reference samples of \CP are drawn over the first $8$ actions of each chunk. In the normalized coordinates of \CP{}, in which every coordinate of the demonstrated actions ranges over $[-1, 1]$, each coordinate of a sample of $\N(0, I_n)$ has a standard deviation equal to half the range of the demonstrated actions, approximately $250$ pixels, and consecutive actions of a sample are independent; consequently, about half of the sampled points lie outside the frame, and the reference of \CP carries no information on the location of the demonstrated actions, in contrast with $P_A$. Only the executed actions are shown, namely $4$ per chunk for \OTP and the first $8$ of $16$ for \CP, colored by the score after their execution. The block at the start (gray) and after step $61$ (dashed) and the target pose (green) are shaded; \CP completes this episode at step $265$.

\paragraph{Figure~\ref{fig:reference}.} In Figure~\ref{fig:reference}, \OTP requires one network evaluation. Each point is the mean over $1{,}500$ episodes, namely three training seeds and $500$ held-out initial conditions, and error bars denote one standard error. The curves show \SIP{}, \FMP{}, and \DPR as functions of $N_\tau$, and \OTP is the single point at $N_\tau = 1$; the vertical axis is logarithmic in $1 - \text{score}$.

\begin{figure}[p]
\centering
\includegraphics[width=\textwidth,height=0.84\textheight,keepaspectratio]{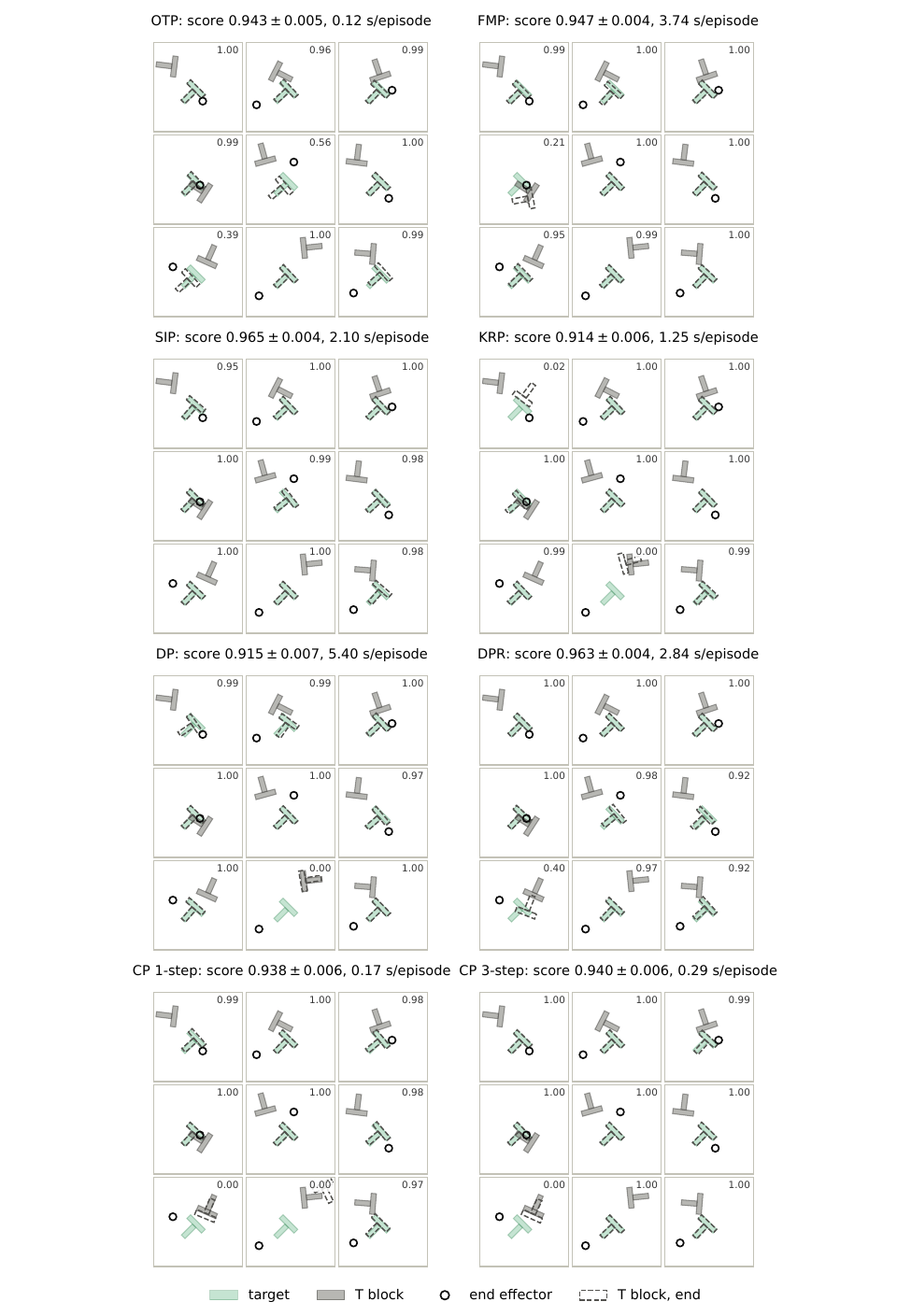}
\caption{Final configurations of nine held-out Push-T episodes, drawn at random and identical for every policy, under the first training seed of each policy: target pose (green), T block at the start (gray), end effector at the start (circle), and T block at the end of the episode (dashed); the number in each scene is the score of the episode. Above each grid: mean score $\pm$ one standard error over the $1{,}500$ episodes of Table~\ref{tab:results}, and the time per episode.}
\label{fig:pushtepisodes}
\end{figure}

\paragraph{Example episodes.} Figure~\ref{fig:pushtepisodes} shows the final configuration of nine held-out episodes, drawn at random from the initial conditions $10000$--$10499$ and identical for every policy, under the first training seed of each policy.

\paragraph{Statistical analysis.} The intervals of Table~\ref{tab:bootstrap} are percentile intervals of $10{,}000$ bootstrap replicates of the difference of the mean scores, in which the $500$ initial conditions are resampled jointly for both policies and the three training seeds are resampled independently for each policy. The seed spread of Table~\ref{tab:results} is the population standard deviation of the three per-seed mean scores; the corresponding sample standard deviations are $0.006$ for \SIP, $0.020$ for \OTP, $0.009$ for \FMP, $0.004$ for \DP, $0.010$ for \DPR, $0.013$ for both variants of \CP, and $0.022$ for \KRP. Since the two variants of \CP share the trained students, the comparison \CP~3-step $-$ \CP~1-step in Table~\ref{tab:bootstrap} resamples the training seeds jointly for both variants.

\begin{table}[t]
\caption{Differences of the mean scores of Table~\ref{tab:results}, with $95\%$ bootstrap intervals.}
\label{tab:bootstrap}
\centering
\footnotesize
\begin{tabular}{@{}lcc@{\hspace{2.5em}}lcc@{}}
\toprule
Comparison & Difference & Interval & Comparison & Difference & Interval \\
\midrule
\SIP $-$ \DP & $0.050$ & $[0.030, 0.072]$ & \FMP $-$ \OTP & $0.004$ & $[-0.021, 0.030]$ \\
\SIP $-$ \OTP & $0.022$ & $[-0.001, 0.046]$ & \OTP $-$ \CP 1-step & $0.005$ & $[-0.023, 0.034]$ \\
\SIP $-$ \FMP & $0.018$ & $[0.001, 0.036]$ & \OTP $-$ \CP 3-step & $0.003$ & $[-0.026, 0.031]$ \\
\SIP $-$ \CP 1-step & $0.027$ & $[0.007, 0.050]$ & \FMP $-$ \CP 1-step & $0.009$ & $[-0.014, 0.034]$ \\
\SIP $-$ \CP 3-step & $0.025$ & $[0.005, 0.047]$ & \FMP $-$ \CP 3-step & $0.007$ & $[-0.016, 0.031]$ \\
\OTP $-$ \DP & $0.028$ & $[-0.001, 0.057]$ & \CP 1-step $-$ \DP & $0.023$ & $[-0.001, 0.044]$ \\
\FMP $-$ \DP & $0.032$ & $[0.009, 0.056]$ & \CP 3-step $-$ \DP & $0.025$ & $[0.003, 0.047]$ \\
\SIP $-$ \KRP & $0.051$ & $[0.024, 0.081]$ & \KRP $-$ \DP & $-0.001$ & $[-0.033, 0.029]$ \\
\OTP $-$ \KRP & $0.029$ & $[-0.004, 0.064]$ & \CP 1-step $-$ \KRP & $0.024$ & $[-0.007, 0.057]$ \\
\FMP $-$ \KRP & $0.033$ & $[0.004, 0.065]$ & \CP 3-step $-$ \KRP & $0.027$ & $[-0.004, 0.060]$ \\
 & & & \CP 3-step $-$ \CP 1-step & $0.003$ & $[-0.006, 0.011]$ \\
\bottomrule
\end{tabular}
\end{table}

\paragraph{Task performance on Push-T.} Table~\ref{tab:results} reports the performance of the seven policies. It is observed that \SIP attains the highest score, $0.965$, followed by \FMP ($0.947$), \OTP ($0.943$), the two variants of \CP ($0.938$ and $0.940$), and \DP ($0.915$). Since episodes that share a training seed or an initial condition are correlated, the differences are assessed by a bootstrap over the initial conditions and the training seeds (Appendix~\ref{app:eval}, Table~\ref{tab:bootstrap}), whose $95\%$ intervals indicate that \SIP exceeds \DP, \FMP, and both variants of \CP, and that \FMP and \CP~3-step exceed \DP, whereas the advantages of \SIP over \OTP and of \OTP and \CP~1-step over \DP are marginal, and \OTP, \FMP, and \CP~1-step are mutually indistinguishable. \KRP, the second policy with a Gaussian reference, attains $0.914$, which is indistinguishable from the score of \DP. \DPR (Diffusion Policy with a residual network), which trains the objective of \DP on the residual network and the integrator of \FMP and \SIP (Appendix~\ref{sec:dpr}), attains $0.963$, and its paired difference from \SIP, $-0.002 \pm 0.005$, is not significant; this suggests that, with $100$ or more network evaluations, the gap between \DP and the proposed policies is largely due to the network and its training rather than to the reference, whose effect is examined in Section~\ref{sec:reference}. The reproduced score of \DP agrees with the average over its last ten checkpoints reported in \citet{chi2023diffusion}, $0.91$ (Appendix~\ref{app:eval}). The policies differ mainly in their failures and in the completion of the task (Appendix~\ref{app:eval}): \DP completes the task as often as \SIP, in $66.7\%$ of the episodes, but fails in $8.6\%$ of them, against $3.1\%$ for \SIP, whereas the deficit of \OTP and \FMP lies in near-completions.

\paragraph{Cost on Push-T.} Table~\ref{tab:cost} reports the offline and online cost of the policies, and Figure~\ref{fig:scoretime} (Appendix~\ref{app:eval}) relates the online cost to the score. On Push-T, the proposed policies are trained in minutes, against $4.5$ hours for \DP and $5.2$ hours for \CP, which reflects both the number of gradient steps and the size of the networks, namely U-Nets of $65.8$ and $69.5$ million parameters against at most $8.1$ million. \DPR, whose network is of the same type as that of \FMP and \SIP ($8.5$ million parameters), is likewise trained in $2.4$ minutes, which indicates that the training cost of \DP is due to its network and its number of gradient steps rather than to its objective. Online, the time per episode follows the number of network evaluations and the size of the network: the one-step policies \OTP and \CP~1-step require $0.12$ and $0.17$ seconds per episode, against $2.1$ to $5.4$ seconds for the policies with $100$ to $200$ evaluations per action chunk, and they attain comparable scores with networks whose sizes differ by a factor of approximately $300$. The two policies differ, however, in the manner in which the single evaluation is obtained: \CP learns its one-step map from a Gaussian reference by distillation, which requires a diffusion teacher of $65.8$ million parameters and a student of $69.5$ million parameters, trained in two stages for $310$ minutes, whereas \OTP learns its one-step map directly, as the conditional OT map from $P_A$, in a single training stage of $10.6$ minutes and with a network of $0.23$ million parameters (Appendix~\ref{app:eval}).

\paragraph{Effect of the reference.} The reference $P_A$ is expected to simplify the learning of the map and to reduce the number of integration steps (Section~\ref{sec:triangular}). To examine this effect, \OTP, \FMP, and \SIP are trained with the reference $P_A$ replaced by $\N(0, I_n)$, every other setting being held fixed, and \FMP and \SIP are deployed with $N_\tau \in \{100, 50, 20, 10, 5, 3\}$ integration steps; the settings were not adapted to the Gaussian reference, and the evaluation protocol is that of Section~\ref{sec:pusht}. Figure~\ref{fig:reference} reports the results, together with those of \DPR (Appendix~\ref{sec:dpr}), a diffusion policy with the network of \FMP and \SIP and the reference $\N(0, I_n)$. With $100$ steps, the two references attain comparable scores, $0.965$ and $0.960$ for \SIP, a paired difference of $0.005 \pm 0.005$, and $0.947$ and $0.955$ for \FMP. As the number of steps is reduced, however, the reference $P_A$ retains the score more effectively: with $3$ steps, \SIP attains $0.798$ with $P_A$ and $0.752$ with $\N(0, I_n)$, and \FMP attains $0.871$ and $0.785$, paired differences of $0.047 \pm 0.008$ and $0.086 \pm 0.007$, whereas the score of \DPR drops to $0.226$ with $5$ steps and to $0.176$ with $3$ steps. For \OTP, which transports the reference with a single evaluation of the map, the effect is decisive: the score drops from $0.943$ with $P_A$ to $0.380$ with $\N(0, I_n)$, a paired difference of $0.563 \pm 0.008$, and the fraction of failed episodes increases from $4.7\%$ to $71.3\%$; with one evaluation, \OTP exceeds \FMP with $3$ steps by $0.072 \pm 0.006$. These results indicate that the reference $P_A$ accounts for a substantial part of the performance of the proposed policies, and that its benefit is largest when the transport is computed with few network evaluations, which is the regime of interest for online deployment.

\paragraph{Comparison of the scores.} In addition to the comparisons of Section~\ref{sec:results}, the intervals of Table~\ref{tab:bootstrap} indicate the following. \KRP, the second policy with a Gaussian reference, attains $0.914$, which is indistinguishable from the score of \DP (difference $-0.001$, interval $[-0.033, 0.029]$); it is exceeded by \SIP ($[0.024, 0.081]$) and by \FMP ($[0.004, 0.065]$), whereas its deficits relative to \OTP and to both variants of \CP are not significant.

\paragraph{Breakdown of the scores.} The breakdown of the scores indicates that the policies differ mainly in their failures and in the completion of the task. \DP completes the task as often as \SIP, in $66.7\%$ of the episodes, but fails, with a score below $0.5$, in $8.6\%$ of them, against $3.1\%$ for \SIP. \OTP and \FMP complete the task in only $51.2\%$ and $52.1\%$ of the episodes but fail in only $4.7\%$ and $4.4\%$ of them; their deficit is concentrated in near-completions, since $35.5\%$ and $34.0\%$ of their episodes attain a score of at least $0.95$ but below one, against $26.5\%$ for \SIP and $22.9\%$ for \DP. \KRP completes the task about as often as \OTP and \FMP, in $55.6\%$ of the episodes, with $31.3\%$ near-completions, but fails in $8.2\%$ of them, a rate close to that of \DP, so that its lower score is accounted for by its failures rather than by its completions.

\paragraph{Computational cost.} In addition to the comparison of Section~\ref{sec:cost}, \KRP is likewise trained in $7.3$ minutes; however, its sampler inverts $\Phi_\theta(s, \cdot)$ one coordinate at a time, with up to $192$ sequential evaluations of a component network per action chunk, so that its time per episode, $1.25$ seconds, lies between those of the one-step policies and those of the policies with $100$ to $200$ evaluations. We remark that \OTP and \CP~1-step are the only policies of Table~\ref{tab:cost} that produce an action chunk with a single network evaluation, and, accordingly, their online times are comparable ($0.12$ and $0.17$ seconds per episode), as are their scores ($0.943$ and $0.938$, a difference of $0.005$ with $95\%$ interval $[-0.023, 0.034]$). The same comparison holds on Square (Section~\ref{sec:square}), where the two policies attain comparable success rates at the same online time, whereas the training of \CP requires $5.3$ hours per seed, against $1.9$ hours for \OTP. Figure~\ref{fig:scoretime} relates the online cost to the score on both tasks. The cost entries of Tables~\ref{tab:results} and~\ref{tab:square} are obtained as follows. Gradient steps: $16{,}384$ iterations for \OTP, each with $25$ updates of the map $T$ and one of the potential $f$; $16{,}384$ for \FMP; $65{,}536$ for \SIP; $32{,}768$ for \KRP; $505{,}000$ for \DP; $150{,}000 + 150{,}000$ for \CP (teacher, then student). The parameter count of \OTP is that of the map $T$; the potential $f$, used in training only, has $2.01$ million parameters. For \KRP, Param.\ counts the weights that the triangular structure leaves active. For \CP, training: teacher, then student ($80 + 230$ minutes on Push-T). With the mean of $8$ chunks, the time per episode on Square changes by at most $0.11$ seconds.

\paragraph{Comparability of published results.} Published results on Push-T differ in the metric, in the observation modality, and in the selection of checkpoints: the metric is the maximum coverage in \citet{chi2023diffusion,prasad2024consistency}, the final intersection over union in \citet{lee2024behavior}, and a success rate in \citet{rana2025imle,gao2026vita}. In particular, \citet{chi2023diffusion} report, for state-based Push-T, $0.95$ as the maximum over checkpoints evaluated on the test initial conditions themselves and $0.91$ as the average over the last ten checkpoints, of which only the latter, which, like the present protocol, involves no selection on the evaluation episodes, is comparable with the scores of Table~\ref{tab:results}, and the state-based results reported for \CP in \citet{prasad2024consistency}, $0.82$ with one step and $0.84$ with three steps, are averages over $200$ episodes of the best checkpoint. Every comparison in this paper is therefore made under the protocol of Section~\ref{sec:pusht}, and a published value is quoted only together with its protocol. 
We note that the released configuration of \citet{chi2023diffusion} for state-based Push-T restricts training to a random subset of $90$ demonstrations, whereas Table~III of \citet{chi2023diffusion} lists $200$ demonstrations for this task; all policies of Table~\ref{tab:results}, \DP included, are trained on the $206$ demonstrations of the data set. 

\paragraph{Replayed demonstrations.} The replayed demonstrations of Table~\ref{tab:results} score $0.947$, below \SIP; since the replay applies the recorded actions in open loop from the recorded initial states, whereas the policies act in closed loop and are scored by the best coverage attained, this value is a property of the data rather than an upper bound on the score of a policy.

\paragraph{Clipping of the executed actions.} Since the executed actions of every policy are clipped (Section~\ref{sec:deployment}), we verified that the clipping is inactive for \DP, whose sampler already confines its output to the demonstrated range: none of the $2{,}504$ actions executed over ten held-out episodes was modified. 

\paragraph{Horizons.} The horizons of the policies differ both in the length of the observation history and in the length of the chunk (Section~\ref{sec:tuning}). 

\section{The Square Task: Settings, Averaging, and State Noise}\label{app:square}

In this appendix, we describe the evaluation on Square (Section~\ref{sec:square}), the diagnosis of the single-sample \OTP on this task, and the two modifications of Table~\ref{tab:square}.

\paragraph{Environment and protocol.} The data set is the low-dimensional proficient-human data set of Square released by \citet{chi2023diffusion}, with absolute actions, and the environment is robosuite~1.2.0 \citep{zhu2020robosuite} through robomimic~0.2.0~\citep{mandlekar2021matters}, the versions with which the data were collected, with the controller set to absolute poses as in the evaluation code of \citet{chi2023diffusion}. The policies receive the $23$-dimensional observation, namely the $14$-dimensional state of the nut and the position ($3$), orientation quaternion ($4$), and gripper joint positions ($2$) of the end effector, and output $10$-dimensional actions, namely the position ($3$), the rotation in the six-dimensional representation ($6$), and the gripper command ($1$); the rotation is converted back to an axis-angle command before it is applied. An episode succeeds if the nut is placed on the peg within $400$ steps, and it stops at the first success. Every entry of Table~\ref{tab:square} uses the final weights of three training runs (seeds $0$, $1$, and $2$), each evaluated on the episode seeds $10000$--$10499$. The two modifications below were selected on the disjoint episode seeds $20000$--$20499$; \DP uses its published configuration for Square, trained for $590{,}000$ gradient steps, with its own normalization. \CP uses the settings of Appendix~\ref{app:baselines}, with the horizons $(2, 16, 8)$, $150{,}000$ gradient steps for each of its teacher and student, and the normalization of \DP{}. The hyperparameters of the proposed policies on Square, with the state noise of Section~\ref{sec:square}, are listed in Table~\ref{tab:hypersquare}.

\begin{table}[t]
\caption{Hyperparameters of the proposed policies on Square.}
\label{tab:hypersquare}
\centering
\footnotesize
\setlength{\tabcolsep}{4pt}
\begin{tabular}{@{}>{\raggedright\arraybackslash}p{0.22\textwidth}
  >{\raggedright\arraybackslash}p{0.27\textwidth}
  >{\raggedright\arraybackslash}p{0.22\textwidth}
  >{\raggedright\arraybackslash}p{0.22\textwidth}@{}}
\toprule
 & \OTP & \FMP & \SIP \\
\midrule
Horizons $(H_o, H_p, H_a)$ & $(2, 4, 4)$ & $(2, 4, 4)$ & $(2, 4, 4)$ \\
Networks & $f$: width $896$, $3$ residual blocks; $T$: width $1344$, $3$ residual blocks & width $1024$, $6$ residual blocks, layer normalization & width $1408$, $6$ residual blocks, layer normalization \\
Activation & mish & mish & mish \\
Time embedding (dimension) & -- & $30$ & $40$ \\
Optimizer & AdamW, weight decay $0$ & AdamW, weight decay $0.001$ & AdamW, weight decay $0.01$ \\
Learning rate & $6.81 \times 10^{-5}$ ($f$), $2.48 \times 10^{-4}$ ($T$) & $3.62 \times 10^{-4}$ & $5.44 \times 10^{-4}$ \\
Schedule & cosine, $5\%$ warm-up & cosine, $5\%$ warm-up & cosine, $5\%$ warm-up \\
Gradient clipping & $1.0$ & $0.5$ & $1.0$ \\
Batch size & $1152$ & $1280$ & $1216$ \\
Iterations & $32{,}768$, with $K_{\mathrm{in}} = 25$ & $131{,}072$ & $131{,}072$ \\
Moving average of the weights (decay) & $0.999$ & $0.9999$ & $0.9999$ \\
Noise level $\varepsilon$ & -- & $0$ & $0.087$ \\
Sampler & one evaluation & Heun, $N_\tau = 100$ & stochastic Heun, $N_\tau = 100$, $\lambda = 1$ \\
\bottomrule
\end{tabular}
\end{table}

\paragraph{Summary of the results.} Table~\ref{tab:square} reports the success rates. Without modification, \OTP succeeds in $76.8\%$ of the episodes, against $91.3\%$ for \SIP and $93.8\%$ for \DP. Unlike Push-T, Square requires the nut to be placed with a precision of a few millimeters, and two deficits of the single-sample \OTP were identified (Appendix~\ref{app:square}): the error of a sampled chunk, about three times that of \SIP, is dominated by its variation with the reference sample, and most failures begin when a small error moves the state away from the demonstrations, where the map extrapolates poorly. Accordingly, we consider two modifications, each directed at one deficit: the coordinate-wise mean of $8$ chunks sampled at the same state, which reduces the variation due to the reference sample, and Gaussian noise added to the state of the robot during training, which exposes the map to states near, rather than on, the demonstrations. It is observed that each modification improves \OTP, to $0.871$ and $0.841$, respectively, and that together they raise its success rate to $0.895$. Since neither modification is specific to \OTP, both are applied to every policy of Table~\ref{tab:square}: the state noise raises the success rate of every policy, whereas the averaging is beneficial only when the variation of the output with the reference sample dominates its error, as is the case for \OTP but not for \SIP. \DPR follows the same pattern: the state noise raises its success rate from $0.896$ to $0.942$, which is comparable with \SIP ($0.936$, a paired difference of $0.006 \pm 0.008$) and \DP ($0.953$, $-0.011 \pm 0.008$), whereas the averaging does not improve it. With both modifications, \OTP is comparable to the other one-step policy, \CP~1-step, in success rate (a difference of $0.005 \pm 0.013$) and in online time, and it trades a deficit of $0.03$ to $0.05$ in success rate, relative to \SIP and \DP, for an order of magnitude in online time (Table~\ref{tab:square}), which the mean of $8$ chunks leaves unchanged, since the $8$ reference samples are mapped in one batched evaluation.

\paragraph{Results.} The observation of Section~\ref{sec:square} comprises the pose of the nut and the position, orientation, and gripper state of the end effector, and the action is the absolute pose command of the end effector, with the rotation in the continuous six-dimensional representation, and the gripper command, as in \citet{chi2023diffusion}. The evaluation protocol is that of Section~\ref{sec:pusht}, namely three training seeds and the same $500$ held-out initial conditions for every policy. Table~\ref{tab:square} reports the success rates. Without modification, \OTP succeeds in $76.8\%$ of the episodes, against $91.3\%$ for \SIP and $93.8\%$ for \DP. Unlike Push-T, Square requires the nut to be placed with a precision of a few millimeters, and two deficits of the single-sample \OTP were identified (see the diagnosis below): the error of a sampled chunk, about three times that of \SIP, is dominated by its variation with the reference sample, and most failures begin when a small error moves the state away from the demonstrations, where the map extrapolates poorly. Accordingly, we consider two modifications, each directed at one deficit: the coordinate-wise mean of $8$ chunks sampled at the same state, which reduces the variation due to the reference sample, and Gaussian noise added to the state of the robot during training, which exposes the map to states near, rather than on, the demonstrations. It is observed that each modification improves \OTP, to $0.871$ and $0.841$, respectively, and that together they raise its success rate to $0.895$, which reduces its deficit relative to \SIP without modification from $0.145$ to $0.019 \pm 0.012$. Since neither modification is specific to \OTP, both are applied to every policy of Table~\ref{tab:square}. For \SIP, the state noise raises the success rate to $0.936$ ($+0.023 \pm 0.010$), whereas the mean of $8$ chunks lowers it to $0.897$ ($-0.017 \pm 0.010$), and, for \DP, the state noise raises it to $0.953$ ($+0.015 \pm 0.007$), whereas the mean of $8$ leaves it unchanged ($0.939$); for \KRP, the state noise raises it from $0.753$ to $0.821$ ($+0.068 \pm 0.014$), whereas the mean of $8$ lowers it, to $0.721$ without and $0.761$ with the state noise. This suggests that the state noise is beneficial beyond \OTP, whereas the averaging is beneficial only when the variation of the output with the reference sample dominates its error, as is the case for \OTP but not for \SIP, whose samples vary about three times less. Table~\ref{tab:square} reports the corresponding costs: \OTP requires $0.12$ seconds per episode, about ten times less than \SIP and \DP ($1.24$ seconds), and the mean of $8$ chunks leaves this time unchanged, since the $8$ reference samples are mapped in a single batched evaluation of the network. For \CP, the state noise raises the success rate to $0.889$ and $0.897$ for one and three steps, respectively, and the mean of $8$ chunks is ineffective without it ($0.880$ and $0.881$), whereas, combined with it, it yields $0.900$ and $0.901$. The other one-step policy, \CP~1-step, is therefore comparable to \OTP both in time ($0.12$ seconds) and, with both modifications, in success rate ($0.900$, against $0.895$ for \OTP, a difference of $0.005 \pm 0.013$); however, its two-stage distillation of a U-Net requires $5.3$ hours of training per seed. With both modifications applied to every policy, \OTP trades a deficit of $0.03$ to $0.05$ in success rate, relative to \SIP and \DP, for an order of magnitude in online time. We remark, moreover, that the mean of sampled chunks is a simple remedy: since the proposed policies represent a distribution of action chunks rather than a single chunk, a more sophisticated sampling rule for the online setting, which exploits this distribution at a cost comparable to that of one network evaluation, is a direction worth investigating, and it is the subject of future work. 

\begin{figure}[p]
\centering
\includegraphics[width=\textwidth,height=0.77\textheight,keepaspectratio]{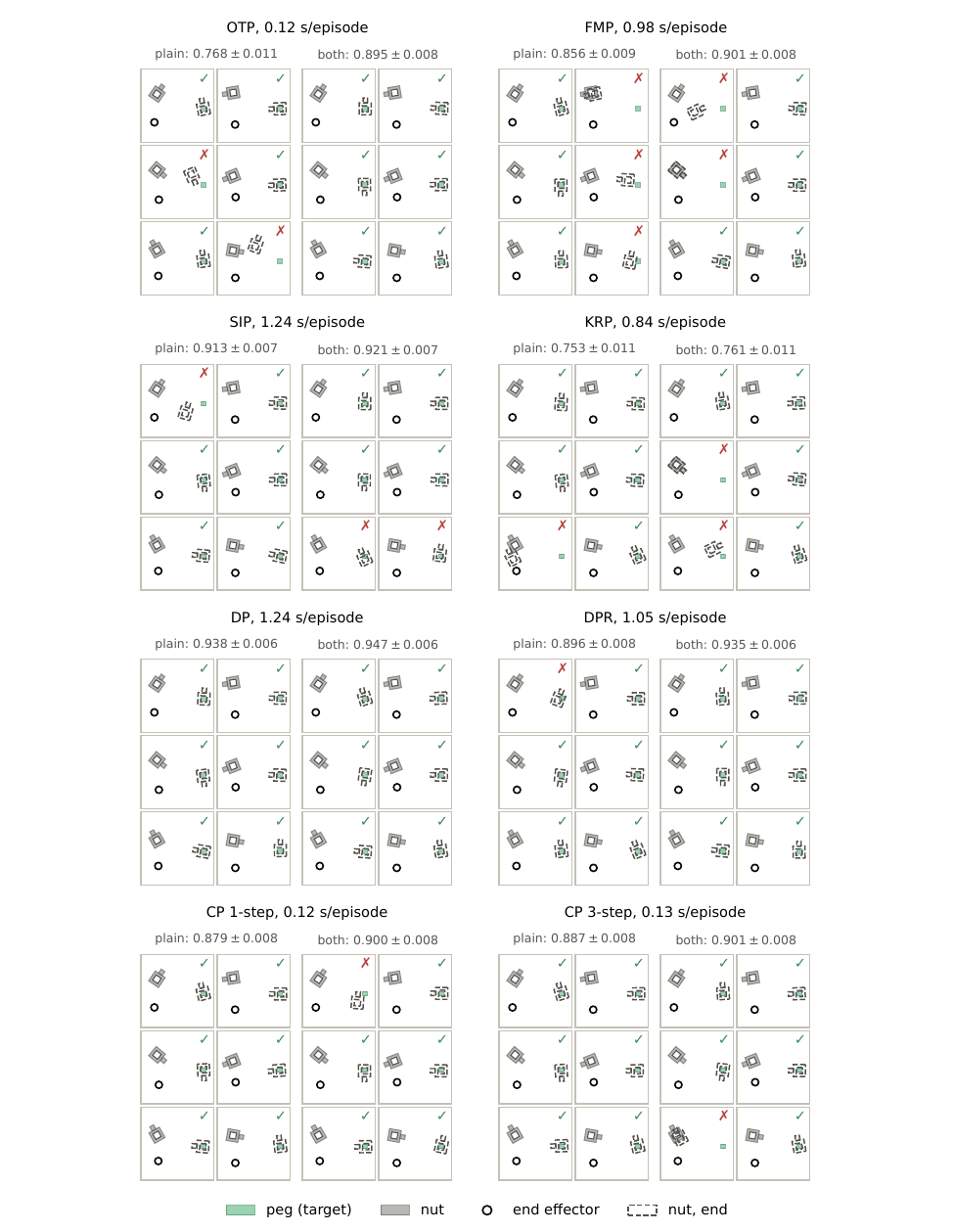}
\caption{Final configurations of six held-out Square episodes, seen from above, drawn at random and identical for every policy and configuration, under the first training seed of each policy; for each policy, the left $3 \times 2$ block is the policy without modification and the right block the policy with both modifications (state noise and mean of $8$ sampled chunks). Each scene shows the square peg (green), the nut at the start (gray), the end effector at the start (circle), and the nut at the end of the episode (dashed); a check mark denotes a success and a cross a failure. Above each block: success rate $\pm$ one standard error over the $1{,}500$ episodes of Table~\ref{tab:square}; above each policy: its time per episode.}
\label{fig:squareepisodes}
\end{figure}

\paragraph{Example episodes.} Figure~\ref{fig:squareepisodes} shows, from above, the final configuration of six held-out episodes, drawn at random from the initial conditions $10000$--$10499$ and identical for every policy and for both configurations, under the first training seed of each policy, without modification (left) and with both modifications (right).

\paragraph{Diagnosis of the single-sample \OTP.} On $256$ demonstrated states, with $64$ reference samples each, the median error of the first commanded position is $15.6$\,mm for \OTP against $4.9$\,mm for \SIP; it consists of the variation of the output across the reference samples at a fixed state, $10.0$\,mm against $3.3$\,mm, and of a bias of the mean output, $12.0$\,mm against $3.9$\,mm, which is largely shared across the training seeds. Across twelve \OTP checkpoints, the single-sample error correlates with the success rate with a coefficient of $-0.94$. In the closed loop, $118$ of the $121$ failed episodes of one training seed contain a jump of the commanded position of more than $10$\,cm between consecutive chunks, against $12$ of its $379$ successful episodes; such a jump typically follows a small error, for instance a grasp that misses the nut, after which the state leaves the region covered by the demonstrations and the map, evaluated outside its training data, produces outputs far outside the demonstrated range. These two observations motivate the two modifications below, which are directed at the variation of the output and at its sensitivity to states near the demonstrations, respectively; neither removes the shared bias.

\paragraph{Mean of sampled chunks.} At every query, $N = 8$ reference chunks $\bar A_k^{(1)}, \ldots, \bar A_k^{(N)}$ are drawn independently, and the executed chunk is
\begin{equation}\label{eq:meanN}
  \hat A_k = \frac{1}{N} \sum_{i=1}^{N} \mathrm{clip}_{[-1, 1]}\bigl( T(S_k, \bar A_k^{(i)}) \bigr),
\end{equation}
computed in normalized coordinates, where each sample is first clipped to the normalized range of the demonstrated actions, so that one extrapolated sample cannot dominate the mean; the gripper command is averaged with the other coordinates, so that its sign is decided by a majority of the samples. For the stochastic policies, the $N$ samples are drawn at the same state with independent noise. The mean~\eqref{eq:meanN} is a Monte Carlo estimate of the mean of the conditional distribution represented by the policy, in the same way as the ensemble mean in the ensemble Kalman filter~\citep{evensen2006}; we are not aware of a policy-learning method that uses it in this form, and it differs from the temporal ensembling of \citet{zhao2023learning}, which averages the chunks predicted at successive time steps for the same future step. Its cost is one batched evaluation of the network with $N$ inputs, which left the time per episode of \OTP unchanged. On the selection block, with three training seeds, the success rate of \OTP was $0.763$ for $N = 1$, $0.823$ for $N = 2$, $0.842$ for $N = 4$, $0.875$ for $N = 8$, $0.866$ for $N = 16$, $0.859$ for $N = 32$, and $0.849$ for $N = 64$, and the coordinate-wise median and trimmed means of $16$ samples did not exceed the mean of $8$; temporal ensembling attained $0.846$ alone and did not improve the mean of $8$. On the evaluation block, the mean of $8$ lowered the success rate of \SIP from $0.913$ to $0.897$ ($-0.017 \pm 0.010$) and left that of \DP unchanged ($0.938$ against $0.939$); since the samples of these policies vary little at a fixed state ($3.3$\,mm for \SIP), the mean removes little error, and it may, moreover, merge distinct chunks at the states at which the demonstrated actions are not unimodal. Since a mean of samples lies between the modes of a multimodal conditional distribution, this modification is suited to the near-unimodal distribution of the demonstrated actions at a given state of Square, and it is not proposed as a general replacement for sampling; a fast sampling rule that exploits the represented distribution more effectively than its mean, for instance by selecting among the samples or by accounting for the previously executed chunk, is left for future work. 

\paragraph{State noise.} During training, the state of every minibatch, on which the policy is conditioned, is replaced by $S + \sigma\, D\, \xi$, with $\xi \sim \N(0, I)$ drawn afresh at every iteration and $\sigma = 0.02$ in the normalized coordinates of \OTP, \FMP, and \SIP, in which each coordinate of the state spans $[-1, 1]$; the diagonal matrix $D$ restricts the noise to the state of the robot, namely the position, quaternion, and gripper joints of the end effector, at both steps of the observation history, and the target actions are left unchanged. In \OTP, the noisy states are also the ones permuted within the minibatch to realize the independent coupling. \DP normalizes the state differently, by a single scale factor, and its standard deviations were therefore set per coordinate so that the noise is the same in physical units. Training with noisy inputs is a known regularizer, equivalent for small noise to a penalty on the derivatives of the model with respect to its input~\citep{bishop1995training}; here, it is applied to the conditioning variable only, so that the map is fitted on states near, rather than on, the demonstrations, which are the states reached after a small error in the closed loop. It is related to the noise injection of \citet{laskey2017dart}, which perturbs the actions of the demonstrator during data collection so that the demonstrations cover such states, whereas the present noise requires no additional demonstrations and leaves the data set unchanged. On the selection block, the success rate of \OTP with the mean of $8$ was $0.892$ for $\sigma = 0.005$ (one training seed), $0.891$ for $\sigma = 0.01$, $0.903$ for $\sigma = 0.02$, and $0.914$ for $\sigma = 0.03$ (one training seed), against $0.875$ without noise, and, with a single sample, $0.803$ for $\sigma = 0.01$ and $0.833$ for $\sigma = 0.02$, against $0.763$; the value $\sigma = 0.02$ was retained on the basis of three training seeds. Applied to \SIP, the same noise raised its success rate on the evaluation block from $0.913$ to $0.936$ ($+0.023 \pm 0.010$, paired over the $1{,}500$ episodes) with a single sample, and to $0.921$ with the mean of $8$.

\section{Extended Discussion}\label{app:discussion}

In this appendix, we give the complete discussion of the results, of which Section~\ref{sec:discussion} retains the conclusions.

\paragraph{Results.} The results of Section~\ref{sec:numerics} indicate that transporting demonstrated action chunks yields policies that are competitive with \DP and \CP on both tasks considered, at a fraction of their computational cost. On Push-T, the three proposed policies attain scores between $0.943$ and $0.965$, against $0.915$ for \DP and $0.938$ to $0.940$ for \CP{}; on Square, with the state noise and the mean of $8$ sampled chunks applied to every policy, they attain success rates between $0.895$ and $0.921$, against $0.947$ for \DP and $0.900$ to $0.901$ for \CP{}. As for the corresponding filters in \citet{aljarrah2026generative}, \OTP is the least expensive and \SIP the most accurate of the proposed policies. In particular, on Push-T, \OTP produces an action chunk with a single evaluation of a network of $0.23$ million parameters, approximately $300$ times fewer than the one-step \CP{}; it is learned in a single training stage of $11$ minutes, without the teacher and the distillation that \CP requires; and its time per episode on Square is approximately one tenth of those of \SIP and \DP{}. These properties make \OTP suited to online settings in which the environment requires an action at every re-planning instant and a delay in its computation cannot be tolerated. We note that these results were obtained with deliberately small networks, of at most $8.1$ million parameters on Push-T and $24.1$ million on Square, against $65.8$ and $65.9$ million for \DP{}, respectively, drawn from a common family of fully connected networks without architectures specific to each policy, since the purpose of the present study is a unified application of the transport methods of \citet{aljarrah2026generative} to imitation learning rather than the attainment of the highest score; the reported scores therefore do not represent the best attainable by the proposed policies, and larger networks and architectures adapted to each policy may improve them further, possibly beyond those of current policies. The advantage of \SIP over \FMP, whose $95\%$ interval $[0.001, 0.036]$ excludes zero only marginally, suggests that the latent noise and the stochastic sampler improve the score, although the two policies use separate settings and differ also in their horizons and networks; similarly, the spread of \OTP across training seeds, three times that of \SIP, suggests a sensitivity to the adversarial training of~\eqref{eq:maxmin}, which an exponential moving average of the weights or a gradient penalty on $f$ may reduce, an attribution that remains a hypothesis. 

\paragraph{Selection of the executed action.} Moreover, the results on Square indicate that the rule by which the executed action is selected deserves attention in its own right. The proposed policies represent a conditional distribution of action chunks rather than a single chunk; if the map is exact, executing one sample reproduces the behavior of the demonstrator, whereas, if the map is approximate, other rules may be preferable, so that the selection of the executed action from the distribution represented by the policy remains, in general, an open question. It is observed that the coordinate-wise mean of $8$ sampled chunks raises the success rate of \OTP from $0.768$ to $0.871$, and to $0.895$ together with the state noise, at an unchanged time per episode, since the $8$ reference samples are mapped in a single batched evaluation of the network (Section~\ref{sec:square}). The same rule, however, lowers the success rate of \SIP and leaves that of \DP unchanged, and, since the mean of chunks drawn from distinct modes need not be an admissible action, it is not expected to be adequate for multimodal conditional distributions. This suggests that the benefit of a selection rule depends on the policy and, in particular, that, for \OTP{}, whose samples are obtained at the cost of a single batched network evaluation, more sophisticated rules may be applied without increasing the online computational time; the design of such rules is the subject of future work.

\paragraph{Limitations and future work.} Several limitations of the present study are noted, each with the corresponding remedy. The comparison is restricted to two state-based tasks, of which Push-T has a conditional action distribution close to unimodal~\citep{mazza2026understanding}, a restriction that tasks with verified conditional multimodality would address; we note that the common alternatives, Block Pushing~\citep{florence2022implicit,shafiullah2022behavior} and Franka Kitchen~\citep{gupta2020relay}, are also estimated to be close to unimodal by \citet{mazza2026understanding}; the baselines are used with their published settings, whereas the settings of the proposed policies were chosen for the present study within the restricted network sizes noted above, an asymmetry that a search over the hyperparameters of the baselines would remove; and the policies are exposed to covariate shift, against which remedies that act on the data set, such as data aggregation when the demonstrator can be queried~\citep{ross2011reduction} or noise injection during the demonstrations~\citep{laskey2017dart}, leave the formulation of Section~\ref{sec:triangular} unchanged. Moreover, Proposition~\ref{prop:triangular} provides no guarantee for approximately trained maps or for the closed-loop system, for which the error analysis of triangular transport maps for filtering~\citep{al-jarrah2025error} indicates a direction, and \OTP produces, at a fixed observation history, a distribution supported on at most $M$ demonstrated chunks, so that the dependence of the performance on the size of the data set should be quantified. Future work will focus on these extensions, on fast and more sophisticated sampling rules that exploit, in the closed loop, the distribution of action chunks represented by the policy, in place of the mean of sampled chunks used on Square (Section~\ref{sec:square}), on observations from images, and on the evaluation of the policies on physical systems. 

\end{document}